\documentclass[10pt, letter, onecolumn]{arxiv}

\usepackage{kantlipsum, lipsum}
\usepackage{dm-colors}
\usepackage{amsmath}
\usepackage{pstricks, pst-node}
\usepackage{verbatim}
\usepackage{multirow}
\usepackage{scalerel}
\usepackage{booktabs}
\usepackage{enumitem}
\usepackage{xspace}
\usepackage{bm}
\usepackage{bbm}
\usepackage{mathtools}
\usepackage{soul}
\usepackage{epsfig}
\usepackage{graphicx}
\usepackage{tcolorbox}
\usepackage{subcaption}
\usepackage{amssymb}
\usepackage{colortbl}
\usepackage{csquotes}
\usepackage{etex}
\usepackage{setspace}
\usepackage{svg}
\usepackage{colortbl}
\usepackage{tabularx,ragged2e}
\usepackage{placeins}
\usepackage[symbol]{footmisc}
\usepackage[bibstyle=nature,citestyle=numeric-comp,%
            natbib=true,backend=biber,maxbibnames=99,%
            giveninits=false,sorting=none]{biblatex}
\usepackage{nameref}
\usepackage{varioref}
\usepackage[pagebackref=false,breaklinks=false,%
            colorlinks=true,bookmarks=true,citecolor=ourdarkblue,%
            urlcolor=ourdarkblue,linkcolor=ourdarkblue]{hyperref}
\usepackage[noabbrev,capitalize]{cleveref}
\usepackage{etoc}

\graphicspath{{figures/}}

\title{{Towards Conversational Diagnostic AI}}

\author[$\ast$,1]{Tao Tu}
\author[$\ast$,1]{Anil Palepu}
\author[$\ast$,1]{Mike Schaekermann}
\author[1]{\\Khaled Saab}
\author[1]{Jan Freyberg}
\author[2]{Ryutaro Tanno}
\author[1]{Amy Wang}
\author[1]{Brenna Li}
\author[1]{Mohamed Amin}
\author[2]{\\Nenad Tomasev}
\author[2]{Shekoofeh Azizi}
\author[1]{Karan Singhal}
\author[2]{Yong Cheng}
\author[1]{Le Hou}
\author[2]{Albert Webson}
\author[1]{\\Kavita Kulkarni}
\author[2]{S. Sara Mahdavi}
\author[1]{Christopher Semturs}
\author[1]{\\Juraj Gottweis}
\author[2]{Joelle Barral}
\author[1]{Katherine Chou}
\author[1]{Greg S. Corrado}
\author[1]{Yossi Matias}
\author[$\dagger$,1]{\\Alan Karthikesalingam}
\author[$\dagger$,1]{Vivek Natarajan}

\affil[1]{Google Research, }
\affil[2]{Google DeepMind}

\renewcommand{\correspondingauthor}[1]{$\ast$~Equal contributions. %
                                       $\dagger$~Equal leadership. \\%
                                       $\ddagger$~Corresponding authors: \{taotu, mikeshake, alankarthi, natviv\}@google.com }

\begin{document}

\begin{refsection}

\begin{abstract}

At the heart of medicine lies the physician-patient dialogue, where skillful history-taking paves the way for accurate diagnosis, effective management, and enduring trust. Artificial Intelligence (AI) systems capable of diagnostic dialogue could increase accessibility, consistency, and quality of care. However, approximating clinicians' expertise is an outstanding grand challenge. Here, we introduce AMIE (Articulate Medical Intelligence Explorer), a Large Language Model (LLM) based AI system optimized for diagnostic dialogue.

AMIE uses a novel self-play based simulated environment with automated feedback mechanisms for scaling learning across diverse disease conditions, specialties, and contexts. We designed a framework for evaluating clinically-meaningful axes of performance including history-taking, diagnostic accuracy, management reasoning, communication skills, and empathy. We compared AMIE's performance to that of primary care physicians (PCPs) in a randomized, double-blind crossover study of text-based consultations with validated patient actors in the style of an Objective Structured Clinical Examination (OSCE). The study included 149 case scenarios from clinical providers in Canada, the UK, and India, 20 PCPs for comparison with AMIE, and evaluations by specialist physicians and patient actors. AMIE demonstrated greater diagnostic accuracy and superior performance on 28 of 32 axes according to specialist physicians and 24 of 26 axes according to patient actors. Our research has several limitations and should be interpreted with appropriate caution. Clinicians were limited to unfamiliar synchronous text-chat which permits large-scale LLM-patient interactions but is not representative of usual clinical practice. While further research is required before AMIE could be translated to real-world settings, the results represent a milestone towards conversational diagnostic AI.

\end{abstract}

\maketitle


\section{Introduction}
\label{sec:introduction}

The dialogue between the physician and the patient is fundamental to effective and compassionate care. The medical interview has been termed ``the most powerful, sensitive, and most versatile instrument available to the physician''~\cite{engel1973interviewing}. In some settings, it is believed that 60-80\% of diagnoses are made through clinical history-taking alone~\cite{peterson1992contributions, hampton1975relative, kassirer1983teaching, roshan2000study, sandler1980importance}. The physician-patient dialogue extends beyond history-taking and diagnosis; it is a complex interaction which establishes rapport and trust, serves as a tool for addressing health needs and can empower patients to make informed decisions that account for their preferences, expectations, and concerns~\cite{silverman2016skills}. Clinicians wield considerable skills in clinical history-taking and the wider ``diagnostic dialogue'', but access to this expertise remains episodic and globally scarce~\cite{rennie2014global}.

Recent progress in general-purpose large language models (LLMs)~\cite{openai2023gpt4, google2023palm2, google2023gemini} has shown that artificial intelligence (AI) systems have capabilities to plan, reason, and incorporate relevant context to hold naturalistic conversations. This progress affords an opportunity to rethink the possibilities of AI in medicine towards the development of fully interactive conversational AI. Such medical AI systems would understand clinical language, intelligently acquire information under uncertainty, and engage in natural, diagnostically useful medical conversations with patients and those who care for them. The potential real-world utility of AI systems capable of clinical and diagnostic dialogue is broad, as the development of such capabilities might improve access to diagnostic and prognostic expertise, to improved quality, consistency, availability, and affordability of care, and to help realize better health outcomes (particularly for populations facing healthcare disparities).

\begin{figure}[ht!]
    \centering
    \includegraphics[width=1\textwidth,height=\textheight,keepaspectratio]{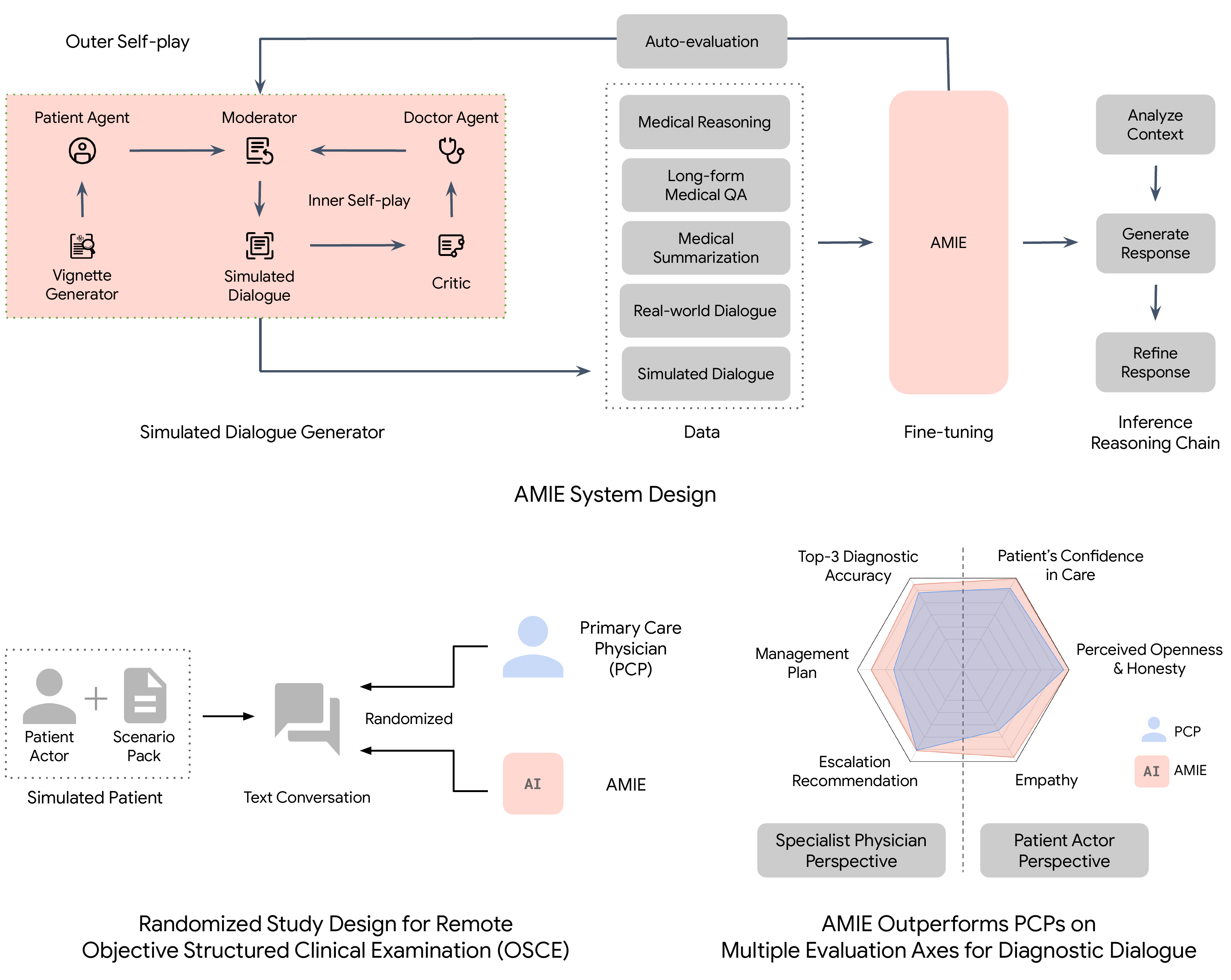}
    \vspace{0.4cm}
    \caption{\textbf{Overview of contributions.} AMIE is a conversational medical AI optimised for diagnostic dialogue. AMIE is instruction fine-tuned with a combination of real-world and simulated medical dialogues, alongside a diverse set of medical reasoning, question answering, and summarization datasets. Notably, we designed a self-play based simulated dialogue environment with automated feedback mechanisms to scale AMIE's capabilities across various medical contexts and specialities. Specifically, this iterative self-improvement process consisted of two self-play loops: (1) An ``inner'' self-play loop, where AMIE leveraged in-context critic feedback to refine its behavior on simulated conversations with an AI patient agent; (2) An ``outer'' self-play loop where the set of refined simulated dialogues were incorporated into subsequent fine-tuning iterations. During online inference, AMIE used a chain-of-reasoning strategy to progressively refine its response conditioned on the current conversation to arrive at an accurate and grounded reply to the patient in each dialogue turn.
    We designed and conducted a blinded remote Objective Structured Clinical Examination (OSCE) with validated simulated patient actors interacting with AMIE or Primary Care Physicians (PCPs) via a text interface. Across multiple axes corresponding to both specialist physician (28 out of 32) and patient actor (24 out of 26) perspective, AMIE was rated as superior to PCPs while being non-inferior on the rest.}
    \label{fig:system_diagram}
\end{figure}

However, while LLMs have been shown to encode clinical knowledge and proven capable of highly accurate single-turn medical question-answering~\cite{singhal2022large,singhal2023towards,nori2023can}, their conversational capabilities have been tailored to domains outside clinical medicine~\cite{thoppilan2022lamda, openai2022chatgpt}. Prior work in LLMs for health ~\cite{singhal2022large, singhal2023towards, nori2023can, toma2023clinical, chen2023meditron} has not yet rigorously examined the clinical history-taking and diagnostic dialogue capabilities of AI systems or contextualized this by comparison to the extensive capabilities of expert clinicians.

Clinical history-taking and diagnostic dialogue through which clinicians derive diagnosis and management plans represent a complex skill~\cite{levine2017history} whose optimal conduct is highly dependent on context. Thus, multiple evaluation axes are needed to assess the quality of a diagnostic dialogue, including the structure and completeness of the elicited history, diagnostic accuracy, the appropriateness of management plans and their rationale, and patient-centred considerations such as relationship-building, respect for the individual and communication efficacy~\cite{king2013best}. If the conversational potential of LLMs is to be realized in medicine, there is a significant unmet need to better optimize development and evaluation of medical AI systems for characteristics such as these, which are unique to history-taking and diagnostic dialogue between clinicians and patients.

In this work, we detail our progress towards a conversational medical AI system for clinical history-taking and diagnostic reasoning.

Our key contributions are summarized as:
\begin{itemize}[leftmargin=1.5em,rightmargin=0em]
\setlength\itemsep{5pt}
\item We introduced AMIE (Articulate Medical Intelligence Explorer), an LLM based AI system optimized for clinical history-taking and diagnostic dialogue.
\item To scale AMIE across a multitude of specialties and scenarios, we developed a novel self-play based simulated diagnostic dialogue environment with automated feedback mechanisms to enrich and accelerate its learning process. We also introduced an inference time chain-of-reasoning strategy to improve AMIE's diagnostic accuracy and conversation quality.  
\item We developed a pilot evaluation rubric to assess the history-taking, diagnostic reasoning, communication skills and empathy of diagnostic conversational medical AI, encompassing both clinician-centred and patient-centred metrics.
\item We designed and conducted a blinded remote OSCE study with 149 case scenarios from clinical providers in Canada, the UK, and India, enabling randomized and counterbalanced comparison of AMIE to PCPs when performing consultations with validated patient actors. AMIE exhibited superior diagnostic accuracy compared to PCPs as assessed by various measures (e.g., top-1 and top-3 accuracy of the differential diagnosis list). Across 28 out of 32 evaluation axes from the specialist physician perspective and 24 out of 26 evaluation axes from the patient actor perspective, AMIE was rated superior to PCPs while being non-inferior on the rest.
\item We performed a range of ablations to further understand and characterize the capabilities of AMIE, highlighted important limitations, and proposed key next steps for real-world clinical translation of AMIE. 

\end{itemize}

Our research has important limitations, most notably that we utilized a text-chat interface, which although enabling potentially large-scale interaction between patients and LLMs specialized for diagnostic dialogue, was unfamiliar to PCPs for remote consultation. Thus our study should not be regarded as representative of usual practice in (tele)medicine.

\begin{figure}[ht]
    \centering
    \includegraphics[width=\textwidth,height=\textheight,keepaspectratio]{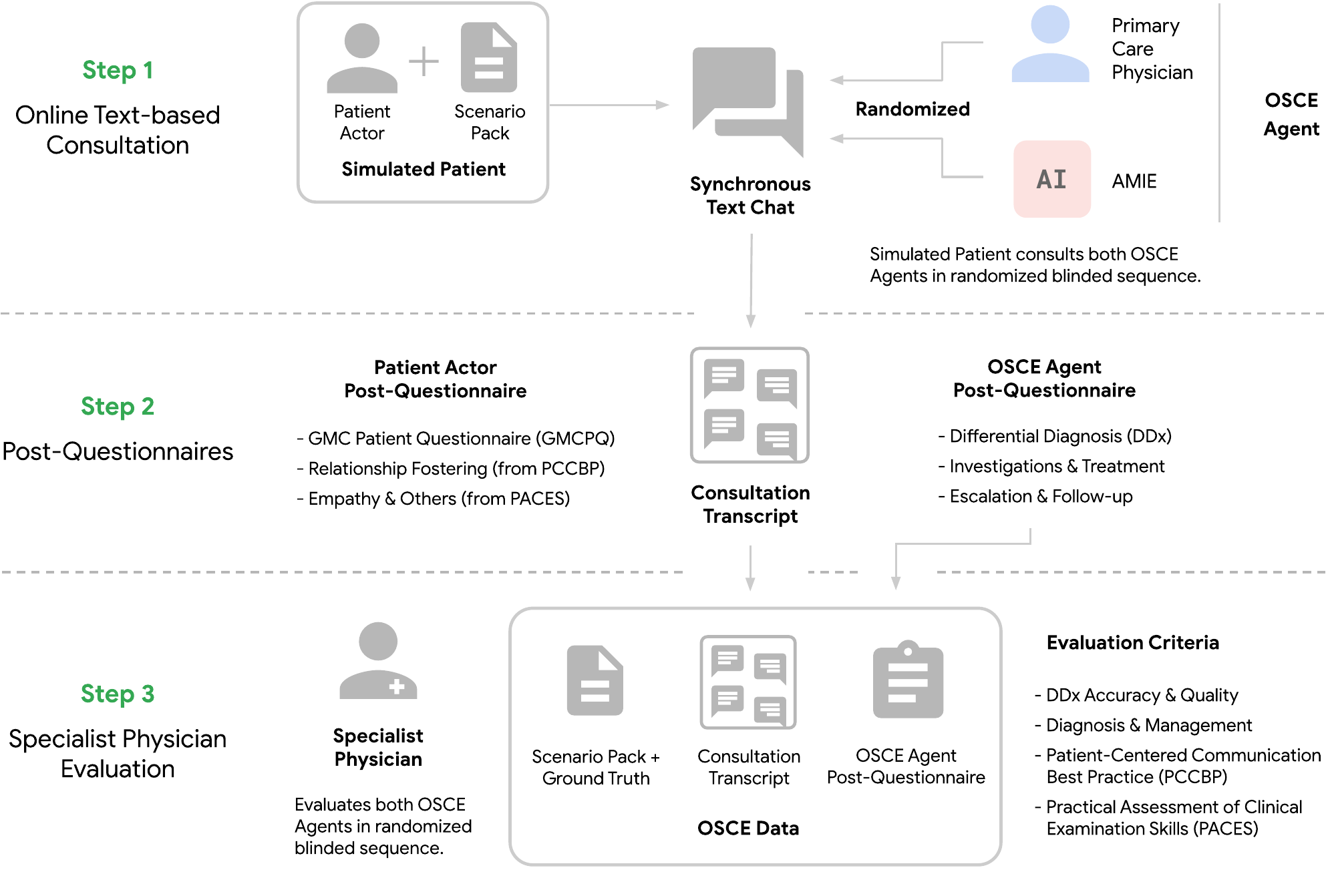}
    \caption{\textbf{Overview of randomized study design.} A primary care physician (PCP) and AMIE perform (in a randomized order) a virtual remote Objective Structured Clinical Examination (OSCE) with simulated patients via online multi-turn synchronous text chat and produce answers to a post-questionnaire. Both the PCP and AMIE are then evaluated by both the patient actors as well as specialist physicians.}
    \label{fig:study_design}
\end{figure}

\newpage

\section{AMIE: An LLM based AI System for Diagnostic Dialogue}
\label{sec:methods}

In the following sections, we describe the real-world datasets, simulated self-play environment, fine-tuning process, and inference time chain-of-reasoning that we designed to optimize AMIE for diagnostic conversation capabilities and clinical communication skills.

\subsection{Real-world Datasets for AMIE}
\label{sec:datasets}

AMIE was developed using a diverse suite of real-world datasets including multiple-choice medical question-answering, expert-curated long-form medical reasoning, electronic health record (EHR) note summaries, and large-scale transcribed medical conversation interactions. As described in detail below, in addition to dialogue generation tasks, the training task mixture for AMIE consisted of medical question-answering, reasoning, and summarization tasks.

\paragraph{Medical Reasoning.} We used the MedQA (multiple-choice) dataset consisting of US Medical Licensing Examination (USMLE) multiple-choice style open domain questions with four or five possible answers~\cite{jin2021disease}. The training set consisted of 11,450 questions and the test set had 1,273 questions. We also curated 191 MedQA questions from the training set where clinical experts crafted step-by-step reasoning leading to the correct answer~\cite{singhal2023towards}.

\paragraph{Long-form Medical Question Answering.} The dataset used here consisted of expert-crafted long-form responses to 64 questions from HealthSearchQA, LiveQA, and Medication QA in MultiMedBench~\cite{singhal2022large}.

\paragraph{Medical Summarization.} A dataset consisting of 65 clinician-written summaries of medical notes from MIMIC-III, a large, publicly available database containing medical records of intensive care unit patients~\cite{johnson2016mimic}, was used as additional training data for AMIE. MIMIC-III contains approximately 2 million notes spanning 13 types including cardiology, respiratory, radiology, physician, general, discharge, case management, consult, nursing, pharmacy, nutrition, rehabilitation and social work. 5 notes from each category were selected, with a minimum total length of 400 tokens and at least one nursing note per patient. Clinicians were instructed to write abstractive summaries of individual medical notes, capturing key information while also permitting the inclusion of new informative and clarifying phrases and sentences not present in the original note.

\paragraph {Real-world Dialogue.} Here, we used a de-identified dataset licensed from a dialogue research organisation comprising 98,919 audio transcripts of medical conversations during in-person clinical visits from over 1,000 clinicians over a 10-year period in the United States~\cite{chiu2017speech}. It covered 51 medical specialties (primary care, rheumatology, hematology, oncology, internal medicine and psychiatry among others) and 168 medical conditions and visit reasons (type II diabetes, rheumatoid arthritis, asthma, depression among the common conditions). Audio transcripts contained utterances from different speaker roles such as doctors, patients, and nurses. On average a conversation had 149.8 turns ($P_{0.25}=75.0$, $P_{0.75}=196.0$). For each conversation, the metadata contained information about patient demographics, reason for the visit (follow-up for pre-existing condition, acute needs, annual exam and more), and diagnosis type (new, existing or other unrelated). We refer to~\citep{chiu2017speech} for more details.

For this study, we selected dialogues involving only doctors and patients, but not other roles such as nurses. During preprocessing, we removed paraverbal annotations such as ``[LAUGHING]'' and ``[INAUDIBLE]'' from the transcripts. We then divided the dataset into training (90\%) and validation (10\%) sets using stratified sampling based on condition categories and reasons for visits, resulting in 89,027 conversations for training and 9,892 for validation.

\subsection{Simulated Dialogue Learning Environment and Self-play for AMIE}

While passively collecting and transcribing real-world dialogues from in-person clinical visits is feasible, two substantial challenges limit its effectiveness in training LLMs for medical conversations: (1) existing real-world data often fails to capture the vast range of medical conditions and scenarios, hindering its scalability and comprehensiveness; (2) the data derived from real-world dialogue transcripts tends to be noisy, containing ambiguous language (including slang, jargon, and sarcasm), interruptions, ungrammatical utterances, and implicit references. This in turn, may limit  AMIE's knowledge, capabilities, and applicability.

To address these limitations, we designed a self-play based simulated learning environment for diagnostic medical dialogues in a virtual care setting, enabling us to scale AMIE's knowledge and capabilities across a multitude of medical conditions and contexts. We used this environment to iteratively fine-tune AMIE with an evolving set of simulated dialogues in addition to the static corpus of medical QA, reasoning, summarization, and real-world dialogue data described above (see \cref{fig:system_diagram}). 

This process consisted of two self-play loops:
\begin{itemize}
    \item \textbf{An ``inner'' self-play loop} where AMIE leveraged in-context critic feedback to refine its behavior on simulated conversations with an AI patient agent.
    \item \textbf{An ``outer'' self-play loop} where the set of refined simulated dialogues were incorporated into subsequent fine-tuning iterations. The resulting new version of AMIE could then participate in the inner loop again, creating a continuous learning cycle. 
\end{itemize} 

\newpage

\renewcommand{\thefootnote}{\arabic{footnote}} \paragraph{Simulated Dialogues.} At each iteration of fine-tuning, we produced 11,686 dialogues, stemming from 5,230 different medical conditions. Conditions were selected from three datasets:
\begin{itemize}
    \item \textbf{Health QA dataset}~\cite{singhal2022large} which contained 613 common medical conditions.
    \item \textbf{MalaCards Human Disease Database}\footnote{\url{https://github.com/Shivanshu-Gupta/web-scrapers/blob/master/medical_ner/malacards-diseases.json}} which contained 18,455 less common disease conditions.
    \item \textbf{MedicineNet Diseases \& Conditions Index}\footnote{\url{https://github.com/Shivanshu-Gupta/web-scrapers/blob/master/medical_ner/medicinenet-diseases.json}} which contained 4,617 less common conditions.
\end{itemize}

At each self-play iteration, four conversations were generated from each of the 613 common conditions, while two conversations were generated from each of the 4,617 less common conditions randomly chosen from
MedicineNet and MalaCards. The average simulated dialogue conversation length was 21.28 turns ($P_{0.25}=19.0$, $P_{0.75}=25.0$).

Using simulated dialogues allowed us to address the limited availability of high-quality, labelled real-world conversation data and improved the model's generalization and adaptability to diverse medical contexts. By leveraging this self-play paradigm, AMIE could continuously learn and refine its conversational and diagnostic capabilities during patient interactions.

\subsubsection{Simulated Dialogue Data Curation} 
\label{Simulated data curation}

In order to produce high-quality simulated dialogues at scale, we developed a novel multi-agent framework which comprised three key components:
\begin{itemize}
    \item \textbf{Vignette Generator}: AMIE leverages web searches to craft unique patient vignettes given a specific medical condition.
    \item \textbf{Simulated Dialogue Generator}: Three LLM agents play the roles of patient agent, doctor agent, and moderator, engaging in a turn-by-turn dialogue simulating realistic diagnostic interactions.
    \item \textbf{Self-play Critic}: A fourth LLM agent acts as a critic to give feedback to the doctor agent for self-improvement. Notably, AMIE acted as all agents in this framework. We describe each component in detail below.
\end{itemize}

\paragraph{Vignette Generator.} \label{patient_vignettes} The vignette generator aimed to create varied and realistic patient scenarios at scale, which could be subsequently used as context for generating simulated doctor-patient dialogues thereby allowing AMIE to undergo a training process emulating exposure to a greater number of conditions and patient backgrounds. The patient vignette (scenario) included essential background information such as patient demographics, symptoms, past medical history, past surgical history, past social history, and patient questions, as well as an associated diagnosis and management plan.

For a given condition, patient vignettes were constructed using the following process. First, we retrieved 60 passages (20 each) on the range of demographics, symptoms, and management plans associated with the condition from using an internet search engine. To ensure these passages were relevant to the given condition, we used the general-purpose LLM, PaLM-2~\cite{google2023palm2}, to filter these retrieved passages, removing any passages deemed unrelated to the given condition. We then prompted AMIE to generate plausible patient vignettes aligned with the demographics, symptoms, and management plans retrieved from the filtered passages, by providing a one-shot exemplar to enforce a particular vignette format. The prompts for each of these steps are as follows:

\begin{tcolorbox}
\textbf{Search Retrieval Template}\\
What are the specific \color{blue}patient demographics/symptoms/management plan \color{black}for the condition \color{blue}[Condition]\color{black}?\\

\textbf{Passage Filtering Template}\\
For the clinical condition, \color{blue}[Condition], \color{black}is the following a good description of common \color{blue}demographics/symptoms/management plans \color{black}(Yes/No)? \\
Description: \color{blue}[Retrieved Passage] \color{black}\\
Answer (Yes/No): \\

\textbf{Vignette Generation Template}\\
The following are several passages about the demographics, symptoms, and management plan for a given condition. Generate 2 different patient vignettes consistent with these passages. Follow the format of the given example (just list N/A if a particular field is unavailable). \\\\
Condition:  \color{blue}[Condition] \color{black} \\
Demographic Passages: \color{blue}[Retrieved Demographic Passages] \color{black}\\
Symptoms Passages: \color{blue}[Retrieved Symptom Passages] \color{black}\\ 
Management Plan Passages: \color{blue}[Retrieved Management Plan Passages] \color{black}\\
Example Format: \color{blue}[Oneshot example]\color{black}\\
Patient Vignettes for \color{blue}[Condition]\color{black}:\\
\end{tcolorbox}

\paragraph{Simulated Dialogue Generator.}

Given a patient vignette detailing a specific medical condition, the simulated dialogue generator was designed to simulate a realistic dialogue between a patient and a doctor in an online chat setting where in-person physical examination may not be feasible. 

Three specific LLM agents (patient agent, doctor agent, and moderator), each played by AMIE, were tasked with communicating amongst each other to generate the simulated dialogues. Each agent had distinct instructions. The patient agent embodied the individual experiencing the medical condition outlined in the vignette. Their role involved truthfully responding to the doctor agent's inquiries as well as raising any additional questions or concerns they may have had. The doctor agent played the role of an empathetic clinician seeking to comprehend the patient's medical history within the online chat environment~\cite{sharma2020computational}. Their objective was to formulate questions that could effectively reveal the patient's symptoms and background, leading to an accurate diagnosis and an effective treatment plan. The moderator continually assessed the ongoing dialogue between the patient agent and doctor agent, determining when the conversation had reached a natural conclusion.

The turn-by-turn dialogue simulation started with the doctor agent initiating the conversation: ``Doctor: So, how can I help you today?''. Following this, the patient agent responded, and their answer was incorporated into the ongoing dialogue history. Subsequently, the doctor agent formulated a response based on the updated dialogue history. This response was then appended to the conversation history. The conversation progressed until the moderator detected the dialogue had reached a natural conclusion, when the doctor agent had provided a differential diagnosis, treatment plan, and adequately addressed any remaining patient agent questions, or if either agent initiated a farewell.

\begin{tcolorbox}
\textbf{Patient Agent Instruction:}\\
\color{black}You are a patient chatting with a doctor over an online chat interface. The doctor has never met you before. 
\color{blue}<patient vignette>
\color{black}Respond to the doctor's questions honestly as they interview you, asking any questions that may come up. \\

\textbf{Doctor Agent Instruction:}\\
You are an empathetic clinician asking a patient about their medical history over an online chat interface. You know nothing about the patient in advance.
Respond to the patient with a single-turn response to better understand their history and symptoms. Do not ask more than two questions. If the patient asks a question, be sure to answer it appropriately. \\

\textbf{Moderator Instruction:}\\
\color{black}The following is a conversation between a doctor and a patient:
\color{blue}<dialog> \color{black}
The conversation should only come to an end if the doctor has finished giving the patient a diagnosis and treatment plan and the patient has no questions left. A conversation also comes to an end if the doctor or patient says goodbye.
Question: has the conversation come to an end? Yes or No.
\end{tcolorbox}

\paragraph{Self-play Critic.}
To ensure high-quality dialogues, we implemented a tailored self-play~\cite{fu2023improving} framework specifically for self-improvement of diagnostic conversations. This framework introduced a fourth LLM agent, acting as a ``critic'' which was also played by AMIE and aware of the ground truth diagnosis, to provide in-context feedback to the doctor agent and enhance its performance in subsequent conversations. The critic agent evaluated the doctor agent's responses based on the following criteria:
\begin{itemize}
    \item The doctor agent exhibits empathy and professionalism while addressing the patient agent's latest questions or comments in a concise manner.
    \item The doctor agent avoids asking too many or repetitive questions (about information already acquired), focusing on a maximum of one or two per response.
    \item The responses should not reveal that the doctor agent is an AI chatbot. They should flow naturally, maintain factual accuracy, and facilitate further engagement from the patient.
    \item The doctor agent asks sufficient questions to identify at least two of the most likely differential diagnoses. They further refine their understanding through targeted questions towards the ground truth diagnosis and offer the corresponding treatment.
\end{itemize}

Following the critic's feedback, the doctor agent incorporated the suggestions to improve its responses in subsequent rounds of dialogue with the same patient agent from scratch. Notably, the doctor agent retained access to its previous dialogue history at each new round. This self-improvement process was repeated twice to generate the dialogues used for each iteration of fine-tuning.

\subsection{Instruction Fine-tuning}
\label{sec:instruction-finetuning}

AMIE, built upon the base LLM PaLM 2~\cite{google2023palm2}, was instruction fine-tuned to enhance its capabilities for medical dialogue and reasoning. We refer to the PaLM-2 technical report for more details on the base LLM architecture.

We employed task-specific instructions to fine-tune AMIE in playing either the patient or doctor role within medical dialogues, performing medical question answering and reasoning, and summarizing EHR notes. While the first round of fine-tuning from the base LLM only used the static datasets, subsequent rounds of fine-tuning leveraged the simulated dialogues generated through the self-play inner loop as described in \cref{Simulated data curation}. 

For dialogue generation tasks, AMIE was trained to predict the next conversational turn based on all previous interactions, assuming either the doctor or patient role. When playing the patient agent, AMIE was prompted to reply to the doctor agent's questions about their symptoms, drawing upon information provided in patient scenarios. These scenarios included patient vignettes (see \cref{patient_vignettes}) for simulated dialogues or metadata such as demographics, visit reason, and diagnosis type for the real-world dialogue dataset. In the doctor agent role, AMIE was prompted to act as an empathetic clinician, interviewing patients about their medical history and symptoms to ultimately arrive at an accurate diagnosis. From each dialogue, we sampled on average 3 turns for each the doctor and patient roles as the target turns to predict based on the conversation leading up to that target turn. Target turns were randomly sampled from all turns in the dialogue that had a minimum length of 30 characters.

Similarly, for the EHR note summarization task, AMIE was provided with a clinical note and prompted to generate a summary of the note. Medical reasoning/QA and long-form response generation tasks followed the same setup as in~\cite{singhal2023towards}. Notably, all tasks except dialogue generation and long-form response generation incorporated few-shot (1-5) exemplars in addition to task-specific instructions for additional context.

\subsection{Chain-of-reasoning for Online Inference}
\label{chain-of-reasoning}

To address the core challenge in diagnostic dialogue - effectively acquiring information under uncertainty to enhance diagnostic accuracy and confidence while maintaining positive rapport with the patient - AMIE employed a chain-of-reasoning strategy before generating a response in each dialogue turn. Here, ``chain-of-reasoning'' refers to a series of sequential model calls, each dependent on the outputs of prior steps. Specifically, we used a three-step reasoning process, described as follows:

\begin{enumerate}
    \item \textbf{Analyzing patient information:} Given the current conversation history, AMIE was instructed to 1) summarize the positive and negative symptoms of the patient as well as any relevant medical/family/social history and demographic information, 2) produce a current differential diagnosis, 3) note missing information needed for a more accurate diagnosis and 4) assess confidence in the current differential and highlight its urgency.
    
    \item \textbf{Formulating response and action:} Building upon the conversation history and the output of step 1, AMIE performed the following: 1) Generate a response to the patient's last message and formulate further questions to acquire missing information and refine the differential diagnosis. 2) If necessary, recommend immediate action, such as an emergency room visit. If confident in the diagnosis based on available information, present the differential.
    
    \item \textbf{Refining the response:} AMIE revises its previous output to meet specific criteria based on the conversation history and outputs from earlier steps. The criteria are primarily related to factuality and formatting of the response (e.g., avoid factual inaccuracies on patient facts and unnecessary repetition, show empathy, and display in a clear format).
\end{enumerate}

This chain-of-reasoning strategy enabled AMIE to progressively refine its response conditioned on the current conversation to arrive at an informed and grounded reply.

\section{Evaluation}

Prior works developing models for clinical dialogue have focused on metrics such as the accuracy of note-to-dialogue or dialogue-to-note generations~\cite{abacha2023overview,ionescu2023overview}, or natural language generation metrics such as BLEU or ROUGE scores that fail to capture the clinical quality of a consultation~\cite{he2022dialmed,naseem2022incorporating}. 

In contrast to these prior works we sought to anchor our human evaluation in criteria more commonly used for evaluating the quality of physicians' expertise in history-taking, including their communication skills in consultation. We derived a framework from principles published in reviews of the consensus for best practices for patient-centered communication (PCCBP) in medical interviews~\cite{king2013best}, criteria examined for history-taking skills by the Royal College of Physicians in the UK as part of their Practical Assessment of Clinical Examination Skills (PACES)\footnote{\url{https://www.mrcpuk.org/mrcpuk-examinations/paces/marksheets}}~\cite{dacre2003mrcp}, and criteria proposed by the UK General Medical Council Patient Questionnaire (GMCPQ)\footnote{\url{https://www.ed.ac.uk/sites/default/files/imports/fileManager/patient_questionnaire\%20pdf_48210488.pdf}} for doctors seeking patient feedback as part of professional re-validation\footnote{\url{https://www.gmc-uk.org/registration-and-licensing/managing-your-registration/revalidation/revalidation-resources/collecting-colleague-and-patient-feedback-for-revalidation}}. We iterated upon these criteria to refine items for inclusion and derived pilot scales and instructions for assessment by using focus groups and interviews with clinicians and OSCE examiners based in the UK, Canada, US, and India. 
Our resulting pilot framework enabled assessment from two perspectives: clinician (board-certified physicians) and lay raters (patient actors). The framework included consideration of consultation quality, structure and completeness, the roles, responsibilities, and skills of the interviewer (Tables \ref{tab:gmcpq_rubric_details}, \ref{tab:paces_rubric_details}, \ref{tab:pccbp_rubric_details}, and \ref{tab:diagnosis_management_rubric_details}).

\subsection{Objective Structured Clinical Examination}

Objective Structured Clinical Examination (OSCE) is a practical assessment format used in healthcare to assess clinical skills and competencies in a standardized and objective fashion~\cite{sloan1995objective, carraccio2000objective, epstein2002defining}.
It differs from traditional written or oral exams that focus primarily on theoretical knowledge and instead aims to provide an environment in which the skills of real-world clinical practice might be assessed.

The OSCE is typically divided into multiple stations (often 8-12), each simulating a real-life clinical scenario enacted by standardized patient actors trained to portray specific symptoms or conditions based on pre-defined scenario descriptions.
At each station, students are given specific tasks to perform, such as taking a clinical history, or making a diagnosis.
Each station has a set time limit, ensuring fairness and efficient assessment.
Trained examiners observe students' performance at each station using a pre-defined checklist or marking scheme.
They assess clinical skills like communication, history-taking, physical examination techniques, clinical reasoning, and decision-making.

\subsection{Remote OSCE Study Design}
To compare AMIE's performance to that of real clinicians, we conducted a randomized crossover study of blinded consultations in the style of a remote OSCE.
Our OSCE study involved 20 board-certified primary care physicians (PCPs) and 20 validated patient actors, 10 each from India and Canada, respectively, to partake in online text-based consultations.
PCPs had between 3 and 25 years of post-residency experience (median 7 years).
Patient actors comprised of a mix of medical students, residents, and nurse practitioners with experience in OSCE participation.
We sourced 149 scenario packs from India (75), Canada (60), and the UK (14).

The scenario packs and simulated patients in our study were prepared by two OSCE laboratories (one each in Canada and India), each affiliated to a medical school and with extensive experience in preparing scenario packs and simulated patients for OSCE examinations. UK scenario packs were sourced from the samples provided on the MRCPUK website. Each scenario pack was associated with a ground truth diagnosis and a set of acceptable diagnoses. The scenario packs covered conditions from cardiovascular (29), respiratory (30), gastroenterology (31), neurology (30), urology, obstetric, and gynecology domains (15), and internal medicine (14). Pediatric or psychiatry domains were excluded from this study, as were intensive care or inpatient case management scenarios.

 Indian patient actors played the roles in all India scenario packs and 7 of the 14 UK scenario packs. Canadian patient actors participated in scenario packs for both Canada and the other half of UK-based scenario packs. This assignment process resulted in 149 distinct simulated patients (``scenarios''). Below, we use the term ``OSCE agent'' to refer to the conversational counterpart interviewing the patient actor, i.e., either PCP or AMIE. ~\cref{tab:osce-summary} summarizes the OSCE assignment information across three geographical locations. Each of the 149 simulated patients completed the three-step study flow depicted in Figure \ref{fig:study_design}.

\begin{table}[ht!]
\footnotesize
\centering
\caption{\textbf{OSCE study summary.} Number of scenario packs, patient actors, simulated patients, and primary care physicians (PCPs) in each of the three locations (Canada, India, and the UK) in the remote OSCE study. 20 board-certified PCPs participated in the study as OSCE agents in comparison with AMIE, 10 each from India and Canada. 20 trained patient actors were involved, with 10 each from India and Canada.
Indian patient actors played the roles in both India and UK scenario packs. Canadian patient actors participated in scenario packs for both Canada and the UK. This process resulted in 149 distinct simulated patients.}
\vspace{1em}
\label{tab:osce-summary}
\begin{tabular}{ccccc}
\toprule
Location & \# of Scenario Packs      & \# of Simulated Patients & \# of Patient Actors  & \# of PCPs   \\ \midrule
Canada  & 60    & 67  & 10 & 10     \\
India  & 75    & 82  & 10 & 10   \\
UK  & 14    & 0  & 0 & 0  \\
\textbf{Total} & \textbf{149}    & \textbf{149}  & \textbf{20} & \textbf{20} \\
 \bottomrule
\end{tabular}
\end{table}

\subsubsection{Online Text-based Consultation}

PCPs and patient actors were primed with sample scenarios and instructions, and participated in pilot consultations prior to the study commencing in order to familiarize themselves with the interface and experiment requirements. 

For the experiment, each simulated patient completed two online text-based consultations via a synchronous text chat interface (Figure \ref{fig:chat_interface}), one with a PCP (control) and one with AMIE (intervention).
The ordering of PCP and AMIE was randomized and patient actors were not informed as to which they were talking to in each consultation.
PCPs were located in the same country as patient actors, and were randomly drawn based on availability at the specified time slot for the consultation.
Patient actors role-played the scenario and were instructed to conclude the conversation after no more than 20 minutes. Both OSCE agents were asked (PCPs via study-specific instructions, and AMIE as part of the prompt template) to not reveal their identity, or whether they were human, under any circumstances.

\subsubsection{Post-questionnaires}

Upon conclusion of the consultation, the patient actor and OSCE agent each filled in a post-questionnaire in light of the resulting consultation transcript (Figure \ref{fig:patient_actor_rating_interface}).
The post-questionnaire for patient actors consisted of the complete GMCPQ (Table \ref{tab:gmcpq_rubric_details}), the PACES components for ``Managing Patient Concerns'' and ``Maintaining Patient Welfare'' (Table \ref{tab:paces_rubric_details}), and a checklist representation of the PCCBP category for ``Fostering the Relationship'' (Table \ref{tab:pccbp_rubric_details}).
Responses patient actors provided to the post-questionnaire are referred to as ``patient actor ratings'' below.
The post-questionnaire for the OSCE agent asked for a ranked differential diagnosis (DDx) list with a minimum of 3 and no more than 10 conditions, as well as recommendations for escalation to in-person or video-based consultation, investigations, treatments, management plan, and the need for a follow-up.

\subsubsection{Specialist Physician Evaluation}

Finally, a pool of 23 specialist physicians from India (14), North America (6), and the UK (3) evaluated PCPs and AMIE with respect to the quality of their consultation, and their responses to the post-questionnaire.
During evaluation, specialist physicians also had access to the full scenario pack along with its associated ground truth differential and additional accepted differentials.
All of the data the specialist physicians had access to during evaluation are collectively referred to as ``OSCE data'' below.
Specialist physicians were sourced to match the specialties and geographic regions corresponding to the scenario packs included in our study, and had between 1 and 36 years of post-residency experience (median 5 years).
Each set of OSCE data was evaluated by one specialist physician randomly assigned to match the specialty and geographic region of the underlying scenario (e.g., Canadian pulmonologist evaluated OSCE data from Canada-sourced respiratory medicine scenario).
Each specialist evaluated OSCE data from both PCP and AMIE for a given scenario.
Evaluations for PCP and AMIE were conducted by the same specialist in a randomized and blinded sequence.

Evaluation criteria included the accuracy, appropriateness and comprehensiveness of the provided DDx list, appropriateness of recommendations regarding escalation, investigation, treatment, management plan and follow-up (Table \ref{tab:diagnosis_management_rubric_details}), and all PACES (Table \ref{tab:paces_rubric_details}) and PCCBP (Table \ref{tab:pccbp_rubric_details}) rating items.
We also asked specialist physicians to highlight confabulations in the consultations and questionnaire responses, i.e., text passages that were non-factual or referred to information not provided in the conversation.
Each OSCE scenario pack additionally supplied specialists with scenario-specific clinical information to assist with rating the clinical quality of the consultation, such as the ideal investigation or management plans; or important aspects of the clinical history that would ideally have been elucidated for the highest quality of consultation possible.

\subsection{Auto-evaluation}
In addition to human evaluations, we implemented model-based auto-evaluation methods as economical consistent alternatives to specialist assessments. These techniques were employed to evaluate both dialogue quality and diagnostic accuracy of the OSCE agent.
To establish the validity of our auto-evaluation methods for assessing dialogue quality, we initially focused on a subset of four evaluation axes from the PACES rubric (\cref{tab:paces_rubric_details}) that were assessed by both the patient actors and the specialist physicians. The auto-evaluation, which uses a self-CoT strategy (details described in~\cref{appendix:auto-eval}) with AMIE to rate dialogues, was in good alignment with human raters and comparable to the inter-specialist agreement on these criteria. For the auto-evaluation of differential diagnoses, we leveraged another LLM, Med-PaLM 2~\cite{singhal2023towards} as a surrogate for a specialist rater to grade the predicted diagnoses against the ground truth diagnoses (more details in~\cref{appendix:auto-eval-ddx}). Our auto-evaluation on DDx accuracy showed a similar trend for AMIE and OSCE agents compared to the specialist ratings. Overall, auto-evaluation trends aligned with human ratings for both dialogue quality and diagnostic accuracy.

We also conducted additional auto-evaluation analyses for the following purposes:
\begin{itemize}
    \item To compare the performance of the DDx accuracy derived from AMIE or PCP consultations;
    \item To compare the DDx accuracy between simulated patients performed in Canada and India and determine if there is systematic differences between the two locations;
    \item To isolate the effects of information acquisition and information interpretation by analyzing the DDx accuracy of AMIE when provided the PCP consultation instead of its own;
    \item To evaluate the efficiency of information acquisition between AMIE and PCPs by analyzing the DDx accuracy as the number of conversation turns increases;
    \item To evaluate the benefit of inner-loop self-play on dialogue quality before and after critic feedback.
\end{itemize}

\subsection{Statistical Analysis}

We evaluated the top-k accuracy of the DDx lists generated by AMIE and PCPs across all 149 simulated patients.
Top-k accuracy was defined as the percentage of cases where the correct diagnosis appeared within the top-k positions of the DDx list.
Specifically, a candidate diagnosis was considered a match if the specialist rater marked it as either an exact match with, very close to or closely related to the ground truth diagnosis (or accepted differential).
Statistical significance for DDx accuracy was determined using bootstrap tests~\cite{horowitz2001bootstrap} with 10,000 samples and false discovery rate (FDR) correction~\cite{benjamini1995controlling} across all k.
Statistical significance for patient actor and specialist ratings was determined using Wilcoxon signed-rank tests~\cite{woolson2007wilcoxon} FDR correction.
Cases where either agent received ``Cannot rate / Does not apply'' were excluded from the test.
Results below refer to $p$-values after FDR correction.

\section{Results}
\label{sec:results}

\subsection{Diagnostic Accuracy}

\begin{figure}
    \centering
    \includegraphics[width=\textwidth,height=\textheight,keepaspectratio]{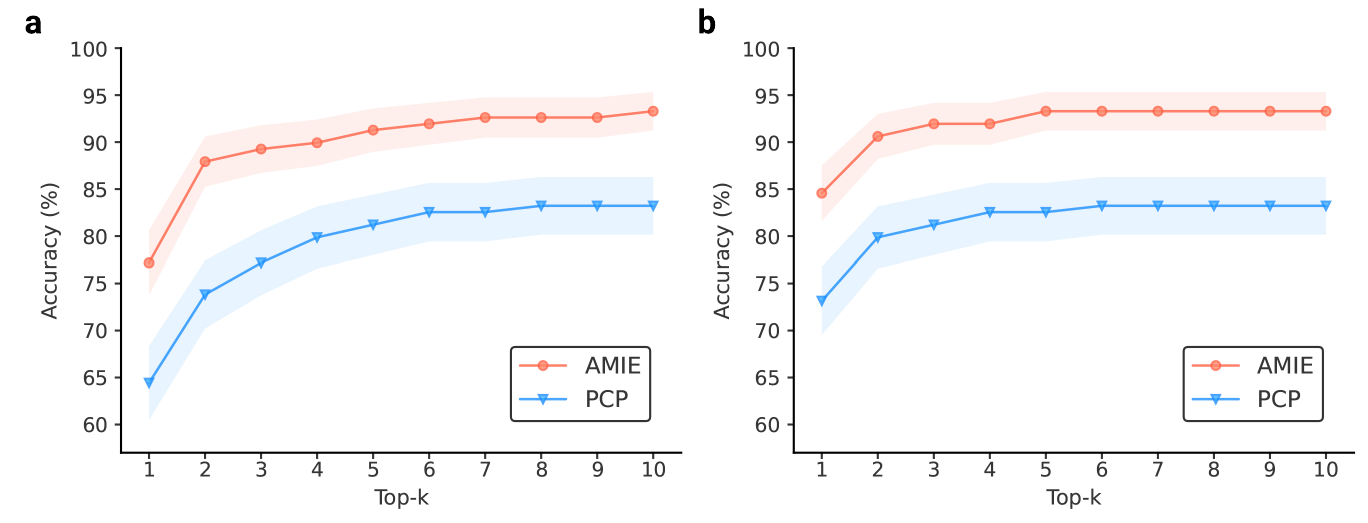}
    \caption{\textbf{Specialist-rated top-k diagnostic accuracy.} AMIE and PCPs top-k DDx accuracy are compared across 149 scenarios with respect to the ground truth diagnosis (\textbf{a}) and all diagnoses in the accepted differential (\textbf{b}). Bootstrapping (n=10,000) confirms all top-k differences between AMIE and PCP DDx accuracy are significant with $p<0.05$ after FDR correction.}
    \label{fig:all_cases_specialist_eval}
\end{figure}

\subsubsection{AMIE showed higher DDx accuracy than PCPs under specialist physician evaluation.}

AMIE's diagnostic accuracy was assessed as higher than that of PCPs.~\cref{fig:all_cases_specialist_eval} shows the top-k accuracy for AMIE and PCPs, considering matches with the ground truth diagnosis (a) and matches with any item on the accepted differential (b). AMIE showed significantly higher top-k accuracy than that of PCPs across all values of k ($p < 0.05$). Note that unlike AMIE, PCPs did not always provide 10 diagnoses in their differential diagnoses (min: 3, mean: 5.39). Additionally, we performed a comparison of DDx accuracy between AMIE and PCP by varying the matching criteria for determining a match. Results depicted in~\cref{fig:all_cases_specialist_match_cutoffs} further substantiate AMIE's superior DDx performance across various matching criteria.

\paragraph{Accuracy by Specialty.}  \cref{fig:specialist_ddx_ratings_by_specialty} illustrates the DDx accuracy achieved by AMIE and PCPs across the six medical specialties covered by scenarios in our study. We observed that AMIE's performance matched or surpassed PCP performance for all specialties with the most pronounced improvements in the respiratory and cardiovascular specialities.

\subsubsection{Auto-evaluation suggested AMIE matched PCPs' efficiency in acquiring information.}

\paragraph{Auto-evaluation Accuracy.}
We reproduced the DDx accuracy analysis with our model-based auto-evaluator instead of the specialist raters using the same procedure as in~\cref{fig:all_cases_specialist_eval}. The overall performance trends obtained through the auto-evaluator align well with specialist assessments despite marginal differences in the computed accuracy values, as shown in~\cref{fig:all_cases_autoeval}.  

\paragraph{Isolating the Source of Performance Gains.} 
To investigate whether AMIE's superior DDx performance observed in~\cref{fig:all_cases_specialist_eval} stemmed from improved information acquisition or from better diagnostic reasoning capability, we compared AMIE's diagnoses based on its own consultations with AMIE's diagnoses generated from the corresponding PCP consultations, using the DDx auto-evaluator. Results depicted in \cref{fig:all_cases_autoeval_AMIEvsAMIE} revealed markedly similar DDx performance, indicating that the diagnostic performance remained consistent regardless of whether AMIE processed information from its own dialogue or from the PCP's conversation. Both methods significantly outperformed the differential diagnoses produced by PCPs. These results suggest that AMIE was approximately equivalent to PCPs at information acquisition but better than PCPs at interpreting that information to produce an accurate/complete differential diagnosis.

\paragraph{Efficiency of Information Acquisition.}

Although AMIE displayed greater verbosity compared to PCPs in terms of total number of words generated in their responses during the consultation, the number of conversational turns and the number of words elicited from the patient actors were similar across both OSCE agents, as illustrated in \cref{fig:number_of_words_and_turns}. This suggests that both AMIE and PCPs acquired a similar amount of information from the patients during the encounter. To investigate how efficient AMIE or PCPs were at gathering sufficient information to formulate a correct diagnosis, we truncated the conversations at various turn counts and used AMIE to generate differential diagnoses based on these partial conversations. ~\cref{fig:all_cases_autoeval_AMIEvsAMIE_turnsablation} depicts the top-3 DDx accuracy as a function of the number of turns provided to the model. The observed accuracies plateaued within the initial 10 conversational turns for both AMIE and PCPs. This suggests that both AMIE and PCPs were able to acquire the information necessary for formulating a diagnosis within the early stages of the conversation. Additionally, the comparable performance at every turn indicates that neither AMIE nor PCPs had a significant advantage in the efficiency or quality of information acquisition.

\begin{figure}[hbtp]
    \centering
    \includegraphics[width=\textwidth,height=\textheight,keepaspectratio]{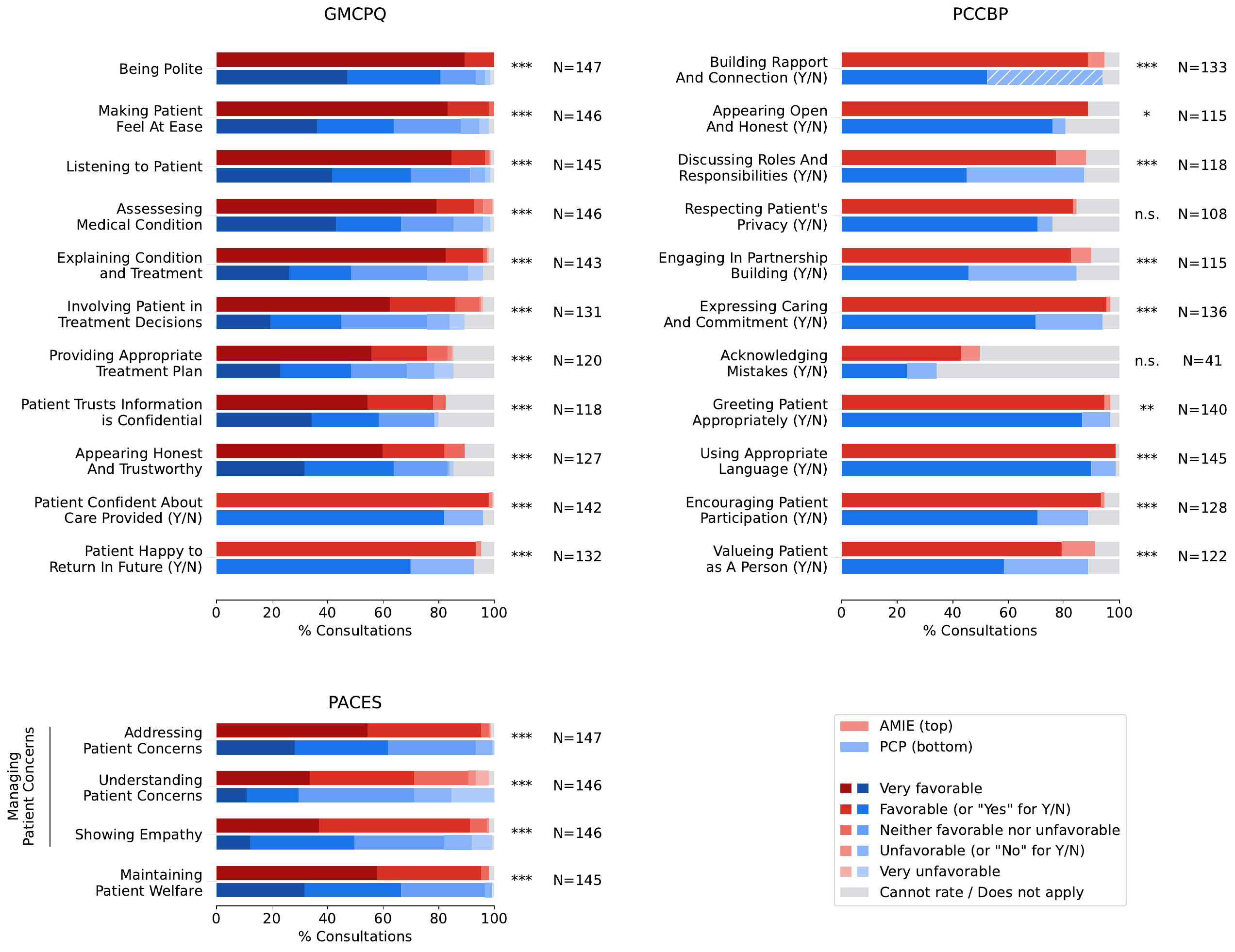}
    \vspace{0.5cm}
    \caption{\textbf{Patient actor ratings.} Conversation qualities as assessed by patient actors upon conclusion of the consultation. For illustration purposes, all responses from five-point rating scales were mapped to a generic five-point scale ranging from `Very favorable' to `Very unfavorable'. For Yes/No questions, a (positive) `Yes' response was mapped to the same color as `Favorable' and a (negative) 'No' response to the same color as `Unfavorable'. Rating scales were adapted from the General Medical Council Patient Questionnaire (GMCPQ), the Practical Assessment of Clinical Examination Skills (PACES), and a narrative review about Patient-Centered Communication Best Practice (PCCBP). Details on question wording and response options are provided in~\cref{appendix:rubrics}. Asterisks represent statistical significance ($*:p<0.05$, $**:p<0.01$, $***:p<0.001$, $n.s.: $ not significant).}
    \label{fig:patient_actor_ratings}
\end{figure}

\subsection{Conversation Quality}

\subsubsection{AMIE surpassed PCPs in conversation quality, per specialists and patient actors.}

Conversation quality was assessed using patient actor ratings, specialist ratings, and outputs from auto-evaluation.
~\cref{fig:AMIE_example_osce} and \ref{fig:pcp_example_osce} show two example consultations for the same simulated patient from AMIE and PCP, respectively.

\paragraph{Patient Actor Ratings.} \cref{fig:patient_actor_ratings} presents the various conversation qualities patient actors assessed following their consultations with the OSCE agents.
Overall, AMIE's consultations were rated significantly better ($p < 0.05$) by patient actors than those from PCPs across 24 of 26 axes. No significant differences in ratings were detected for the two PCCBP axes ``Respecting Patient's Privacy'' (N=108) and ``Acknowledging Mistakes'' (N=41). For the latter criterion, the number of exclusions was substantially higher since the question applied only when mistakes were made by the OSCE agent and pointed out in the conversation.

\begin{figure}[hbtp]
    \centering
    \includegraphics[width=\textwidth, height=\textheight, keepaspectratio]{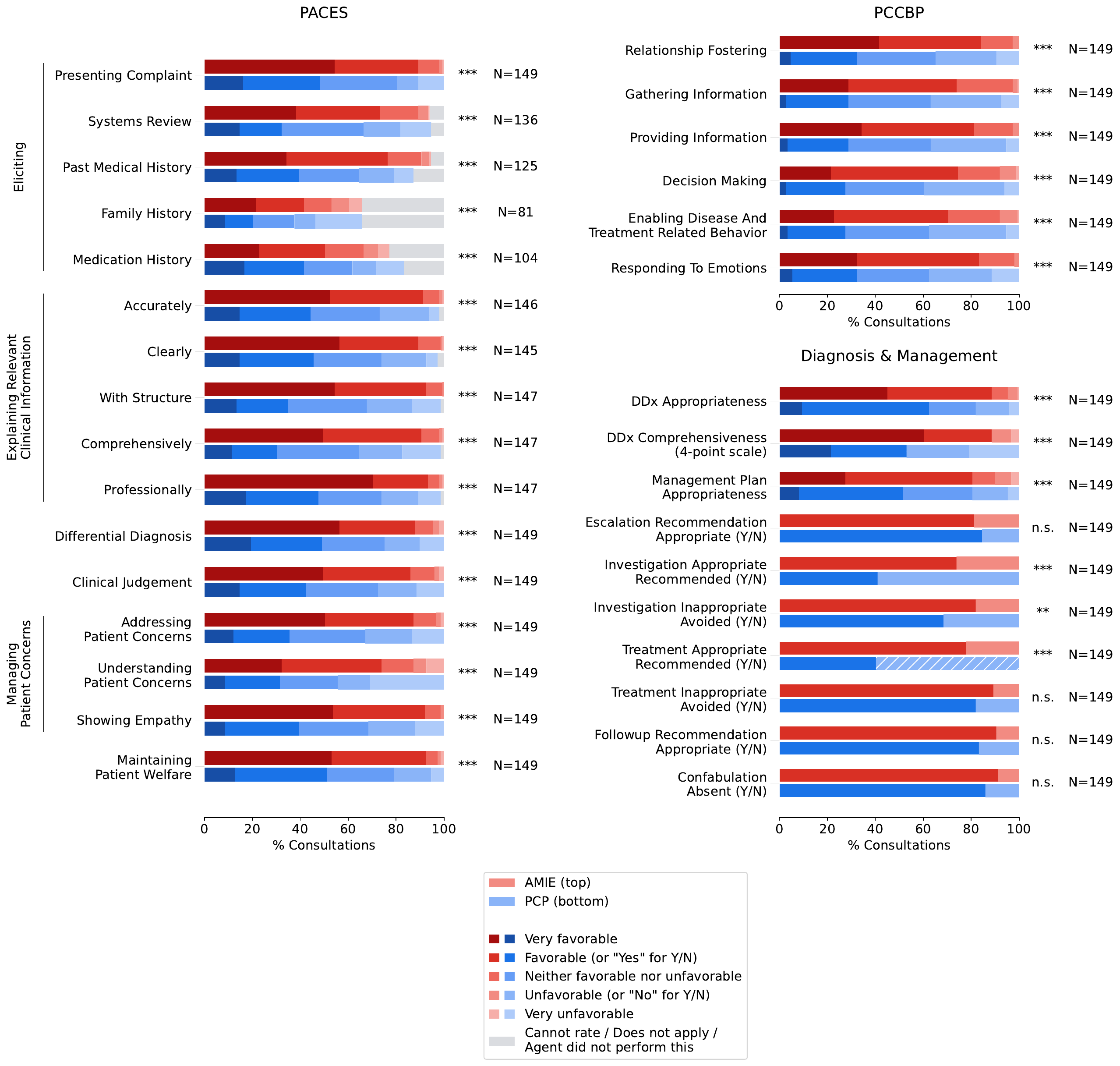}
    \vspace{0.5cm}
    \caption{\textbf{Specialist physician ratings.} Conversation and reasoning qualities as assessed by specialist physicians. For illustration purposes, all responses from five-point rating scales were mapped to a generic five-point scale ranging from `Very favorable' to `Very unfavorable'. The only four-point scale (DDx Comprehensiveness) was mapped to the same scale, ignoring the `Neither favorable nor unfavorable' option. For Yes/No questions, a (positive) `Yes' response was mapped to the same color as `Favorable' and a (negative) 'No' response to the same color as `Unfavorable'. Rating scales were adapted from the Practical Assessment of Clinical Examination Skills (PACES), a narrative review about Patient-Centered Communication Best Practice (PCCBP), and other sources. Details on question wording and response options are provided in~\cref{appendix:rubrics}. Asterisks represent statistical significance ($*:p<0.05$, $**:p<0.01$, $***:p<0.001$, $n.s.: $ not significant).}
    \label{fig:specialist_ratings}
\end{figure}

\paragraph{Specialist Physician Ratings.} Specialist physicians evaluated both the conversational quality as well as the responses to the post-questionnaire for scenarios within their domain expertise (see \cref{fig:specialist_ratings}). Again, AMIE's responses were rated significantly better by specialists than those from PCPs on 28 of 32 evaluation axes; Specialists preferred AMIE's consultation, diagnoses, and management plan over those from PCPs. For this set of evaluations, differences in specialist ratings between AMIE and PCPs were statistically significant ($p < 0.05$).
No significant differences in ratings were detected for four of the axes in the Diagnosis \& Management rubric, namely, ``Escalation Recommendation Appropriate'', ``Treatment Inappropriate Avoided'', ``Followup Recommendation Appropriate'' and ``Confabulation Absent'', despite no exclusions (N=149).

\subsubsection{Auto-evaluations demonstrated the effectiveness of inner self-play for AMIE.}

\paragraph{Auto-evaluation of Conversation Ratings.}
We leveraged the model-based self-CoT auto-evaluation strategy to rate conversations on four evaluation axes from the PACES rubric, and validated that these auto-evaluation ratings were accurate and well aligned with the specialist ratings (\cref{fig:autoeval_ablation,fig:autoeval_vs_specialist}). Furthermore, to demonstrate that the inner self-play loop improved simulated dialogue quality, we applied the auto-evaluation method to the simulated dialogues generated before and after the self-play procedure. Results in~\cref{fig:autoeval_selfplay} revealed that the simulated dialogues after self-play were preferred more often than the baseline dialogues without self-critique.

\section{Related Work}
\label{sec:related-work}

\subsection{Clinical History-taking and the Diagnostic Dialogue}

History-taking and the clinical interview are widely taught in both medical schools' and postgraduate curricula~\cite{keifenheim2015teaching, yedidia2003effect, makoul2003communication, tan2021teaching, raper2015improving, von2008uk}. Consensus on physician-patient communication has evolved to embrace patient-centred communication practices, with recommendations that communication in clinical encounters should address six core functions: fostering the relationship, gathering information, providing information, making decisions, responding to emotions and enabling disease- and treatment-related behavior~\cite{king2013best, de2009endpoints, epstein2007patient}. Specific skills and behaviours for meeting these goals have also been described, taught and assessed~\cite{schirmer2005assessing, king2013best} with validated tools~\cite{schirmer2005assessing}. Medical conventions consistently cite that certain categories of information should be gathered during a clinical interview, comprising topics such as the presenting complaint, past medical history and medication history, social and family history, and systems review~\cite{nichol2018medical, denness2013consultation}. Clinicians' ability to meet these goals is commonly assessed using the framework of an objective structured clinical examination (OSCE)~\cite{sloan1995objective, carraccio2000objective, epstein2002defining}. Such assessments vary in their reproducibility or implementation and have even been adapted for remote practice as virtual OSCEs (vOSCEs) with telemedical scenarios, an issue of particular relevance during the COVID-19 pandemic~\cite{chan2023implementation}.  

\subsection{Conversational AI and Goal-oriented Dialogue}

Conversational AI systems for goal-oriented dialogue and task completion have a rich history~\cite{budzianowski2018multiwoz, wei2018airdialogue, lin2023decisionoriented}. The emergence of transformers~\cite{vaswani2017attention} and large language models~\cite{thoppilan2022lamda} have led to renewed interest in this direction. The development of strategies for alignment~\cite{ouyang2022training}, self-improvement~\cite{zhao2021ethical, saunders2022self, scheurer2023training, glaese2022improving} and scalable oversight mechanisms~\cite{bai2022constitutional} have enabled large scale deployment of such conversational systems in the real world~\cite{openai2022chatgpt, askell2021general}.
However, the rigorous evaluation and exploration of conversational and task-completion capabilities of such AI systems remains limited for clinical applications, where studies have largely focused on single-turn interaction use cases such as question-answering or summarization.

\subsection{AI for Medical Consultations and Diagnostic Dialogue}
The majority of explorations of AI as tools for conducting medical consultations have focused on ``symptom checker'' applications rather than a full natural dialogue, or on topics such as transcription of medical audio or the generation of plausible dialogue given clinical notes or summaries~\cite{shor2023clinical, abacha2017overview, wallace2022diagnostic, zeltzer2023diagnostic}. Language models have been trained using clinical dialogue datasets but not comprehensively evaluated~\cite{johri2023testing}. Studies have been grounded in messages between doctors and patients in commercial chat platforms (which may have altered doctor-patient engagement compared to 1:1 medical consultations)~\cite{zeng2020meddialog, liu2022meddg, he2022dialmed}. Many focused largely on predicting next turns in the recorded exchanges rather than clinically meaningful metrics. And to date, there have been no reported studies that have examined the quality of AI models for diagnostic dialogue using the same criteria that are used to examine and train human physicians in dialogue and communication skills; nor evaluating AI systems in common frameworks such as the OSCE.

\subsection{Evaluation of Diagnostic Dialogue}
Prior frameworks for human evaluation of AI systems' performance in diagnostic dialogue have been limited in detail. They have not been anchored in established criteria for assessing communication skills and the quality of history-taking. For example,~\cite{naseem2022incorporating} reported a 5-point scale describing overall ``human evaluation'',~\cite{zeng2020meddialog} reported ``relevance, informativeness and human likeness'',~\cite{liu2022meddg} reported ``fluency, expertise and relevance'',~\cite{varshney2022cdialog} ``fluency and adequacy'' and~\cite{yan2022remedi} ``fluency''. These criteria are far less comprehensive and specific than those taught and practiced by medical professionals. A multi-agent framework for assessing conversational capabilities of LLMs is introduced in~\cite{johri2023testing}, however, the study was performed in the restricted setting of dermatology, used AI models to emulate both doctor and patient sides of simulated interactions, and performed limited expert evaluation of history-taking as ``complete'' or not.

\section{Discussion}
\label{sec:discussion}

In this study, we introduced AMIE, an LLM based AI system optimised for clinical dialogue with diagnostic reasoning capabilities. We compared AMIE consultations to those performed by PCPs using a randomized, double-blind crossover study with human simulated patients in the style of an Objective Structured Clinical Examination (OSCE). Notably, our study was not designed to be representative of clinical conventions either for traditional OSCE evaluations, for remote- or tele-medical consultation practices, or for the ways clinicians usually use text and chat messaging to communicate with patients. Our evaluation instead mirrored the most common way by which people interact with LLMs today, leveraging a potentially scalable and familiar mechanism for AI systems to engage in remote diagnostic dialogue. In this setting, we observed that AMIE, an AI system optimised specifically for the task, outperformed PCPs on simulated diagnostic conversations when evaluated along multiple clinically-meaningful axes of consultation quality.
 
\paragraph{Diagnostic Performance.} The differential diagnoses provided by AMIE were more accurate and complete than those provided by board-certified PCPs, when both were evaluated by specialist physicians. Previous research has shown that AI systems may match or exceed human diagnostic performance in specific, narrow tasks~\cite{kelly2019key, mcduff2023towards, kanjee2023accuracy} in retrospective evaluation. However, these situations typically involved both AI and physicians interpreting the same fixed input (for example, identifying the presence of a specific finding in a medical image). Our study was significantly more challenging because it required the AI system to actively acquire relevant information through conversation rather than relying on clinical information collated by human efforts~\cite{semigran2015evaluation}. Therefore the system's downstream differential diagnoses depended on not only its diagnostic inference capability, but also the quality of information gathered under uncertainty through natural conversation and building rapport.

Our results suggested that AMIE was as adept as PCPs in eliciting pertinent information during the simulated consultations and was more accurate than PCPs in formulating a complete differential diagnosis if given the same amount of acquired information. This finding corroborates other work that LLMs may be able to produce more complete differential diagnoses given the same clinical information as physicians in challenging cases ~\citep{mcduff2023towards}. Though not explored in this study, the assistive performance of AMIE therefore represents an interesting and important avenue for future research, particularly given the real-world importance of expert oversight for AI systems in safety-critical settings such as medicine.

Our study utilized a wide variety of simulated patients, comprising actors trained in both Canada and India and scenarios across a range of specialties. This allowed us to explore how performance varied along multiple axes: by specialty, and by the locations in which the scenario was derived and enacted. We observed that both PCPs and AMIE performed worse in obstetric/gynecology and internal medicine scenarios than those from other specialties (see~\cref{fig:specialist_ddx_ratings_by_specialty}). The study was not powered or designed to compare performance between different specialty topics, and we cannot exclude that the scenarios in some specialties might be harder than others. We observed that both AMIE and PCPs had higher diagnostic accuracy in consultations performed in the Canada OSCE lab compared to those enacted in the India OSCE lab (see~\cref{fig:per_location_specialist_eval}). However, the differences were not statistically significant and in a subset of 40 scenarios enacted in both the Canada OSCE lab and the India OSCE lab, the performance of both AMIE and PCPs was equivalent (see~\cref{fig:same_scenario_location_ddx}). 

\paragraph{Conversational Performance.} Patient actors and specialist raters both evaluated AMIE's performance to be higher than PCPs on metrics related to empathy and communication skills. These axes comprised a majority of the dimensions that were evaluated. This general finding is consistent with a prior study where LLM responses were found to be more empathetic than the responses from clinicians to health questions posted on Reddit~\citep{ayers2023comparing}. However, the findings in that study may not be generalised directly to our setting due to the differences in study design. Specifically, prior work has not involved a direct, randomised comparison of physicians and AI systems in a prospective simulation of multi-turn dialogue with the same patient. In both settings, the lack of voice-based and non-verbal visual communication may be an unfair disadvantage to clinicians. 

The text-based chat interface used in this study introduces both advantages and disadvantages. People today most commonly engage with LLMs through synchronous text-chat interfaces~\cite{openai2023website}, and patients often use patient portals to send messages to their providers. We therefore chose this mode of interaction as a representative interface for LLMs to perform multi-turn conversation, adapting the virtual OSCE framework accordingly. While this allowed a fair comparison of diagnostic dialogue between LLMs and clinicians when both were restricted to a synchronous text-chat, it is important to acknowledge that our experiments do not emulate the expected quality of diagnostic dialogue in real clinical practice (including telemedicine). Physicians may be more used to history-taking and diagnostic dialogue by telephone or video consultation than synchronous text-chat communication~\cite{carrillo2022effectiveness, wharton2019virtual}. Instead, text is more commonly used by clinicians to communicate with patients for episodic or asynchronous needs such as prescription refills or communication about specific test results~\cite{fuster2022asynchronous}. Physicians may thus be more familiar with text/SMS or email rather than the synchronous text-chat medium we employed in this study. In both text/SMS and email, the conventions and expectations for communicating naturally and with empathic style might be different~\cite{hammersley2019comparing}. It is possible that the PCPs in our study had not yet become accustomed to the setting, and may have performed differently if subjected to a specific training program (similar in spirit to the training process for AMIE). Clinicians participating in the study undertook two preparatory pilot sessions of consultations with our synchronous text interface before the evaluation began, but this was not a formal training program, nor was it designed to optimize clinicians' performance. Future research could explore this question more thoroughly including monitoring for the impact of a learning curve, or exploring whether performance varies according to the extent to which participating clinicians or simulated patients are familiar with telemedicine.

Additionally, our findings regarding empathic communication could also be partially attributed to the fact that AMIE responses were significantly longer than clinician responses (shown in~\cref{fig:number_of_words_and_turns}), and presented with greater structure. This could potentially suggest to an observer that more time was spent preparing the response, analogous to known findings that patient satisfaction increases with time spend with their physicians~\citep{gross1998patient, tates2017effect, zyzanski1998trade}.

Collectively, our findings suggest many avenues for further research that might leverage human-AI complementarity~\cite{dvijotham2023enhancing}, combining clinicians' skills in the analysis of verbal and non-verbal cues with the potential strengths of LLMs to suggest more enriched conversational responses including empathic statements, structure, eloquence, or more complete differential diagnoses.

\paragraph{Simulated Dialogue.} The use of simulated data allowed us to quickly scale training to a broad set of conditions and patient contexts, while the injection of knowledge from search encouraged these dialogues to remain grounded and realistic. Though the simulated patients encompassed a wide range of conditions, they failed to capture the full range of potential patient backgrounds, personalities, and motivations. Through the inner self-play procedure, we were able to iteratively improve the simulated dialogue we generated and used in fine-tuning. However, these improvements were limited by our ability to articulate what makes a good dialogue in the critic instructions, the critic's ability to produce effective feedback, and AMIE's ability to adapt to such feedback. For example, in the simulated environment we impose that AMIE reaches a proposed differential and testing/treatment plan for the patient, but such an endpoint may be unrealistic for some conditions, especially in the virtual chat-based setting.

\paragraph{Evaluation Framework.} In contrast to prior works, we anchored our evaluation in criteria already established to be relevant for assessing physicians' communication skills and history-taking quality. We performed more extensive and diverse human evaluation than prior studies of AI systems, with ratings from both clinicians and simulated patients perspective. Our raters and scenarios were sourced from multiple geographic locations, including North America, India and the UK. Our pilot evaluation rubric is, to our knowledge, the first to evaluate LLMs' history-taking and communication skills using axes that are also measured in the real world for physicians themselves, increasing the clinical relevance of our research. Our evaluation framework is considerably more granular and specific than prior works on AI-generated clinical dialogue, which have not considered patient-centred communication best practice or clinically-relevant axes of consultation quality~\cite{naseem2022incorporating, zeng2020meddialog, liu2022meddg, varshney2022cdialog, yan2022remedi, johri2023testing}. 

However, our pilot framework is not definitive and can be further improved in future research. History-taking itself is contextual and what determines a ``good history'' is dependent on the specific clinical situation, patient and physician attributes, cultural characteristics, and many other factors. Despite variation in models for clinical history-taking~\cite{bird1990three,rezler1991missing,rosenberg1997lessons,smith2002patient}, studies have shown that good clinical interviews are associated with not only problem detection and diagnostic accuracy, but also quadruple aims for care delivery~\cite{berwick2008triple,bodenheimer2014triple} ranging from patient and physician satisfaction, resilience to stress and illness, and health outcomes or cost. 
Future studies on the quality of LLM history-taking might therefore utilise prospective measures of these outcomes in real-world settings (for example reductions in patient complaints~\cite{adamson1989physician}, or improvements in cost and care effectiveness, patient and provider satisfaction), though evaluations as such may be challenging or impractical to compare to standard practice in the same individual patient, and randomisation of different approaches may also be challenging in real-world settings.

\paragraph{Breadth of Evaluation.} Our chosen axes of evaluation were not exhaustive and their interpretation was often subjective in nature. Although we conducted evaluations from both clinician and lay-perspectives, generating scenario-packs in three countries with assessors in both North America and India, the pool of clinicians and lay-people assessing the models could be expanded further to improve generalization of our insights. Our experiments could also undergo more extensive replication to explore other aspects such as inter-observer and inter-participant variability, including future work with an intentionally further diversified pool of human raters (clinicians and lay users). Participatory design in the development of model evaluation tools with a representative pool of patients, as well as clinical and health equity domain experts, could also be valuable.

Although our scenarios comprised many different clinical conditions and specialties, our experiments were not necessarily representative of the decades of clinical practice accumulated by even a single doctor (who on average may perform tens of thousands of consultations in a career~\cite{silverman2010doctors}). The range of conditions possible to examine in medicine is vast as is the variation in presentation of individual diseases. Our experiments were not designed to examine multi-morbidity and co-incident pathology, longitudinal case presentation or the consideration of sequential information from clinical investigations. We excluded entirely some clinical settings or specialties such as psychiatry, pediatrics, intensive care, and inpatient case management scenarios. Further research would be needed to understand the applicability of our findings in many settings such as these, where the requirements for high-quality history-taking might differ~\cite{rahman2023inter,kantar2022history}. The OSCE framework is commonly used in the assessment of clinicians' skills. It encompasses a significant range of methodologies including real or simulated patients, interaction with physical artefacts or clinical materials, applications to a variety of medical specialties, tasks or settings; and both remote or in-person assessments. Although the OSCE approach is popular, there are significant limitations to its validity~\cite{setyonugroho2015reliability}. We utilised a remote text-based assessment, replicating known issues with the paradigm of ``virtual OSCE'' such as the inability to incorporate non-verbal symptoms, signs and communication features. Additionally, this format could introduce unfamiliar constraints to the communication of PCP participants ~\cite{chan2023implementation}. 

The tone, content, and nature of the OSCE dialogues in our study are likely not to be representative of real-world patient populations. For example, patient actors may have described their symptoms with greater structure, depth or clinical detail than could be routinely expected in many consultations, or had greater comprehension of clinical context than would be ordinarily expected. Furthermore, although evaluation was blinded, the style of responses from AMIE was notably different to that by PCPs which limits the practical extent of blinding in study design.
 
Therefore even within the distribution of diseases and specialties we addressed, our findings should be interpreted with humility and caution. There is a need for further research to examine varied presentations of the same diseases, alongside exploration of alternate approaches to evaluating history-taking and clinical dialogue in situations of different patient needs, preferences, behaviours and circumstances.

\paragraph{Fairness and Bias.} The evaluation protocol presented in this paper is limited in terms of its ability to capture potential issues related to fairness and bias, which remains an important open question that we will aim to address in subsequent system evaluations. Recent advances in the development of comprehensive frameworks for bias detection in large language models~\cite{weidinger2022taxonomy, gallegos2023bias} present a promising starting point for establishing such an approach. It should be noted that medical diagnostic dialogue is a particularly challenging use case, due to the complexity of the medical domain, the interactive information gathering nature of the dialogue, and the outcome-driven setting, with the potential of associated harms in case of incorrect diagnosis or incorrect medical advice. Nevertheless, disentangling these issues is an important further research area if LLMs in the domain are to overcome rather than propagate inequities in healthcare. For example, previous studies have found that physicians approach communication with their patients differently, on average, depending on patients’ race, resulting in Black patients receiving communication that was less patient-centered, and with a lower positive affect~\cite{johnson2004patient}. Other studies have found differences in physicians’ communication styles and conversation length based on gender~\cite{roter2002physician}. Effective intercultural communication skills are essential~\cite{rahman2023inter}. There is therefore a non-negligible risk that such historical conversational biases may be replicated or amplified in an AI dialogue system, but at the same time there is also an opportunity to work towards designing conversational systems that can be more inclusive, and more personalized to the individual patient’s needs.

To help inform the development of the necessary fairness, bias, and equity frameworks, it is important to employ a participatory approach to solicit representative views across a wide range of patient demographics, as well as clinical and health equity domain experts. Such evaluation frameworks should be complemented by extensive model red teaming and an adversarial approach to identifying any remaining gaps and failure modes. Recent advances in red teaming LLMs could be useful in this scenario~\cite{perez2022red, ganguli2022red, yu2023gptfuzzer, ge2023mart}. These practices should not only inform the evaluation of the final model, but also its development and iterative refinement. Model development should follow the established data and model reporting practices and provide transparency into the training data and the associated decision processes~\cite{mitchell2019model, crisan2022interactive, pushkarna2022data}. The dialogue research dataset contributing to AMIE training data in our study was de-identified, reducing the availability of socio-economic factors, patient demographics, and information about clinical settings and locations. 

Further work is also needed to ensure the robustness of medical LLMs in multilingual settings~\cite{choudhury2021linguistically, talat2022you, ahuja2023megaverse, ImaniGooghari_2023}, and particularly their performance in low-resource languages~\cite{nguyen2023democratizing}. The great variety of cultures~\cite{naous2023having}, languages, localities, identities, and localized medical needs, makes the task of generating a priori static yet comprehensive fairness benchmarks practically infeasible. Measurement and mitigation of bias must move beyond the traditional narrow focus on specific axes that fails to scale globally~\cite{ramesh2023fairness}. LLM-based evaluators present a potential solution for preliminary assessments in languages where there are no systematic benchmarks, though prior studies have found these auto-evaluation frameworks to be biased, underscoring the need for calibrating them on native speaker evaluations, and using them with caution~\cite{hada2023large}.

\paragraph{Deployment.} This research demonstrates the potential of LLMs for future use in healthcare in the context of diagnostic dialogue. Transitioning from an LLM research prototype that has been evaluated in this study to a safe and robust tool that can be used by healthcare providers, administrators, and people will require significant additional research to ensure the safety, reliability, efficacy, and privacy of the technology. Careful consideration will need to be given to the ethical deployment of this technology including rigorous quality assessment across different clinical settings and research into reliable uncertainty estimation methods~\cite{quach2023conformal, chen2023quantifying, huang2023look, yang2023uncertaintyaware} that would allow for deferral to human clinical experts when needed. These and other guardrails are needed to mitigate potential overreliance on LLM technologies, with other specific measures for attention to ethical and regulatory requirements particular to future use-cases and the presence of qualified physicians in the loop to safeguard any model outputs. Additional research will also be needed to assess the extent to which biases and security vulnerabilities might arise either from base models or the circumstances of use in deployment, as we have highlighted in our prior work~\cite{singhal2022large}. Given the continuous evolution of clinical knowledge, it will also be important to develop ways for LLMs to utilize up-to-date clinical information~\cite{lazaridou2021mind}.

\section{Conclusion}
\label{sec:conclusion}
The utility of medical AI systems could be greatly improved if they are better able to interact conversationally, anchoring on large-scale medical knowledge while communicating with appropriate levels of empathy and trust. 
This research demonstrates the significant potential capabilities of LLM based AI systems for settings involving clinical history-taking and diagnostic dialogue. The performance of AMIE in simulated consultations represents a milestone for the field, as it was assessed along an evaluation framework that considered multiple clinically-relevant axes for conversational diagnostic medical AI. However, the results should be interpreted with appropriate caution. Translating from this limited scope of experimental simulated history-taking and diagnostic dialogue, towards real-world tools for people and those who provide care for them, requires significant additional research and development to ensure the safety, reliability, fairness, efficacy, and privacy of the technology. If successful, we believe AI systems such as AMIE can be at the core of next generation learning health systems that help scale world class healthcare to everyone.

\subsubsection*{Acknowledgments}
This project was an extensive collaboration between many teams at Google Research and Google DeepMind.
We thank Yun Liu, Daniel McDuff, Jake Sunshine, Ali Connell, Paul McGovern and Zoubin Ghahramani for their comprehensive review and detailed feedback on the manuscript. We also thank Sami Lachgar, Lauren Winer, John Guilyard and Maggie Shiels for contributions to the narratives and visuals.
We are grateful to Julie Anne Seguin, Sally Goldman, Yuri Vasilevski, Xinying Song, Akshay Goel, Chu-ling Ko, Abhinav Das, Haiyang Yu, Chang Liu, Yuchen Liu, SiWai Man, Brett Hatfield, Sean Li, Ajay Joshi, Gordon Turner, Annisah Um'rani, Divya Pandya and Preeti Singh  for their valuable insights, technical support and
feedback during our research.
We also thank our clinical provider partners in Canada and India for their partnership in conducting the OSCE study. Finally, we are grateful to Dale Webster, Ewa Dominowska, David Fleet, Philip Mansfield, Sushant Prakash, Renee Wong, Susan Thomas, Michael Howell, Karen DeSalvo, Jeff Dean, James Manyika,  Zoubin Ghahramani and Demis Hassabis for their support during the course of this project.

\subsubsection*{Data Availability}
Some of the real-world datasets used in the development of AMIE are open-source (MedQA). The scenario packs from UK used in the OSCE study are also available for download on the internet. 

\subsubsection*{Code Availability} AMIE is an LLM based research AI system for diagnostic dialogue. We are not open-sourcing model code and weights due to the safety implications of unmonitored use of such a system in medical settings. In the interest of responsible innovation, we will be working with research partners, regulators, and providers to validate and explore safe onward uses of AMIE. For reproducibility, we have documented technical deep learning methods while keeping the paper accessible to a clinical and general scientific audience.  Our work builds upon PaLM 2, for which technical details have been described extensively in the technical report \cite{google2023palm2}.

\subsubsection*{Competing Interests}
This study was funded by Alphabet Inc and/or a subsidiary thereof (‘Alphabet’). All authors are employees of Alphabet and may own stock as part of the standard compensation package. 

\newpage
\setlength\bibitemsep{3pt}
\printbibliography
\balance
\clearpage

\end{refsection}

\newpage
\begin{refsection}

\clearpage

\renewcommand{\thesection}{A.\arabic{section}}
\renewcommand{\thefigure}{A.\arabic{figure}}
\renewcommand{\thetable}{A.\arabic{table}} 
\renewcommand{\theequation}{A.\arabic{equation}} 
\renewcommand{\theHsection}{A\arabic{section}}

\setcounter{section}{0}
\setcounter{figure}{0}
\setcounter{table}{0}
\setcounter{equation}{0}


\noindent \textbf{\LARGE{Appendix}}\\
\normalfont

In the following sections, we report additional data and detailed analyses to further illustrate the performance of AMIE.

We provide details on:
\begin{itemize}[leftmargin=1.5em,rightmargin=0em]
\item OSCE evaluation rubrics 

\item Example of simulated dialogues after self-critique

\item AMIE user interfaces

\item Example consultation with OSCE agents
\item DDx top-k accuracy by degree of matching
\item DDx top-k accuracy by specialty
\item Auto-evaluation on DDx
\begin{itemize}
    \item Reproducing DDx accuracy via auto-evaluation
    \item AMIE DDx accuracy on AMIE and PCP consultations
    \item DDx accuracy as a function of the number of dialogue turns
 \end{itemize}
\item DDx top-k accuracy by location
\item  Model-based auto-evaluation of qualitative criteria:
\begin{itemize}
    \item Rank-order agreement of auto-evaluation to specialists
    \item Auto-evaluation of simulated dialogues with self-play
 \end{itemize}

\end{itemize}

\vspace{0.8cm}

\section{OSCE Evaluation Rubrics}
\label{appendix:rubrics}

\begin{table}[hbt!]
    \centering
    \includegraphics[width=\textwidth,height=\textheight,keepaspectratio]{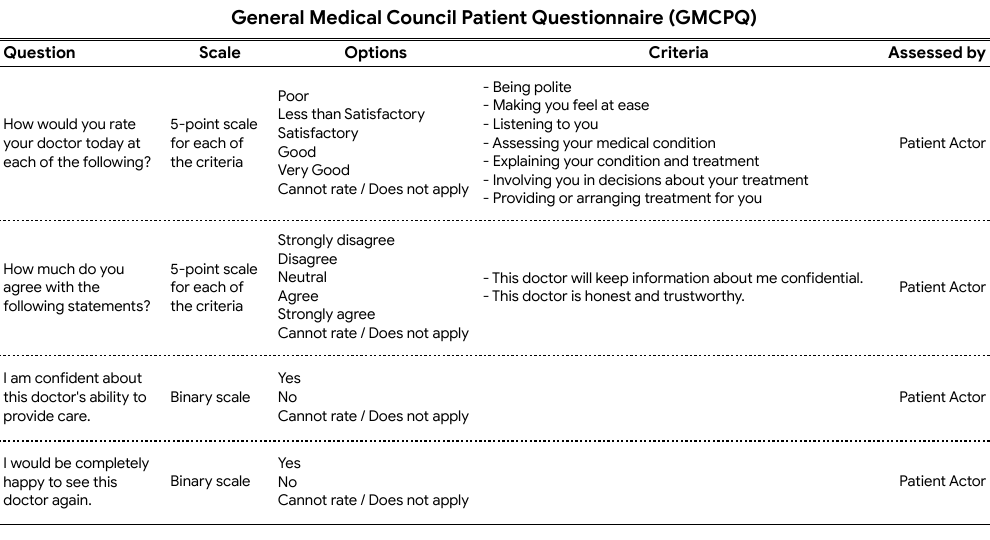}
    \caption{\textbf{General Medical Council Patient Questionnaire (GMCPQ) rubric details.}}
    \label{tab:gmcpq_rubric_details}
\end{table}

\begin{table}[hbt!]
    \centering
    \includegraphics[width=\textwidth,height=\textheight,keepaspectratio]{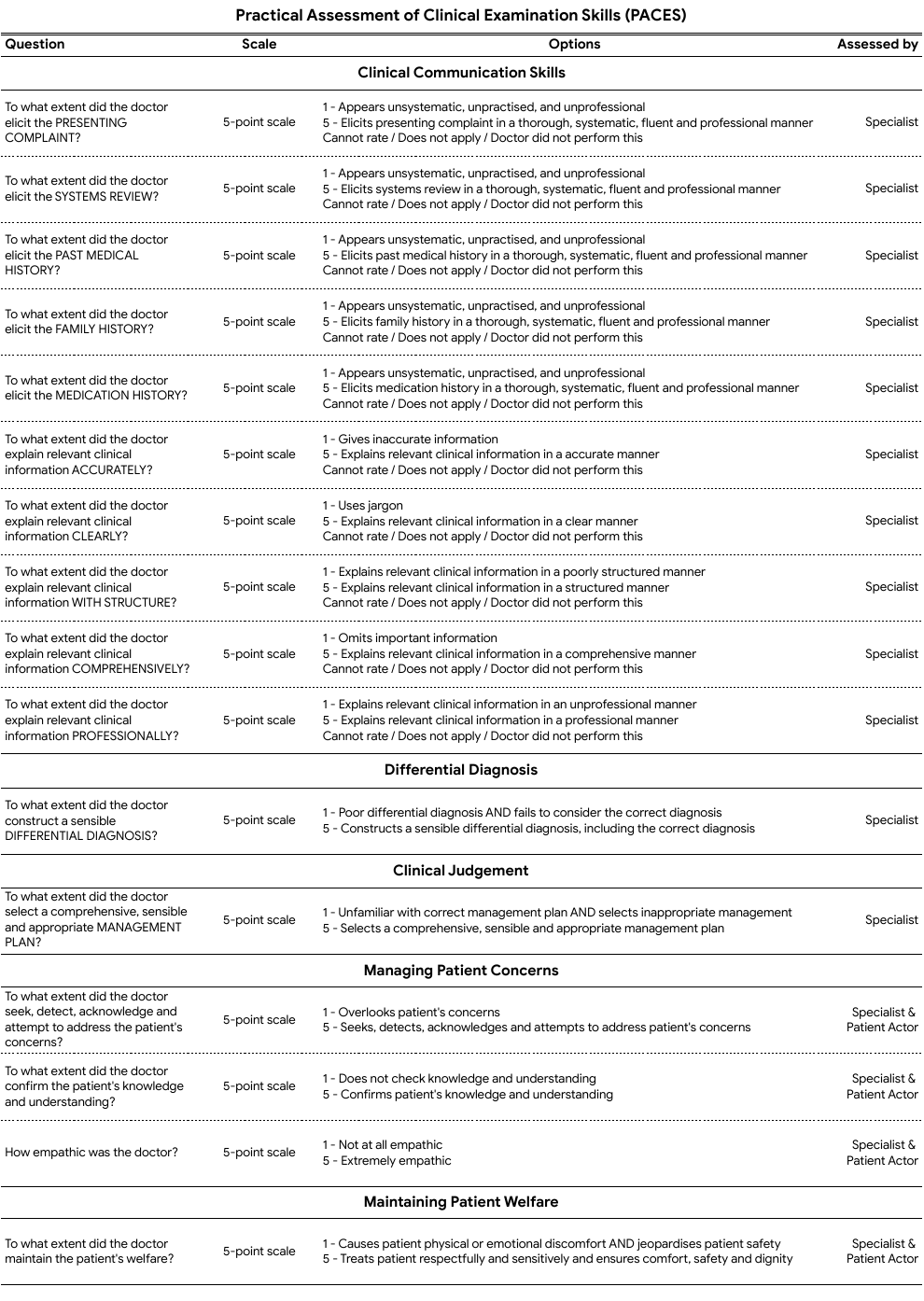}
    \caption{\textbf{Practical Assessment of Clinical Examination Skills (PACES) rubric details.}}
    \label{tab:paces_rubric_details}
\end{table}

\begin{table}[hbt!]
    \centering
    \includegraphics[width=\textwidth,height=\textheight,keepaspectratio]{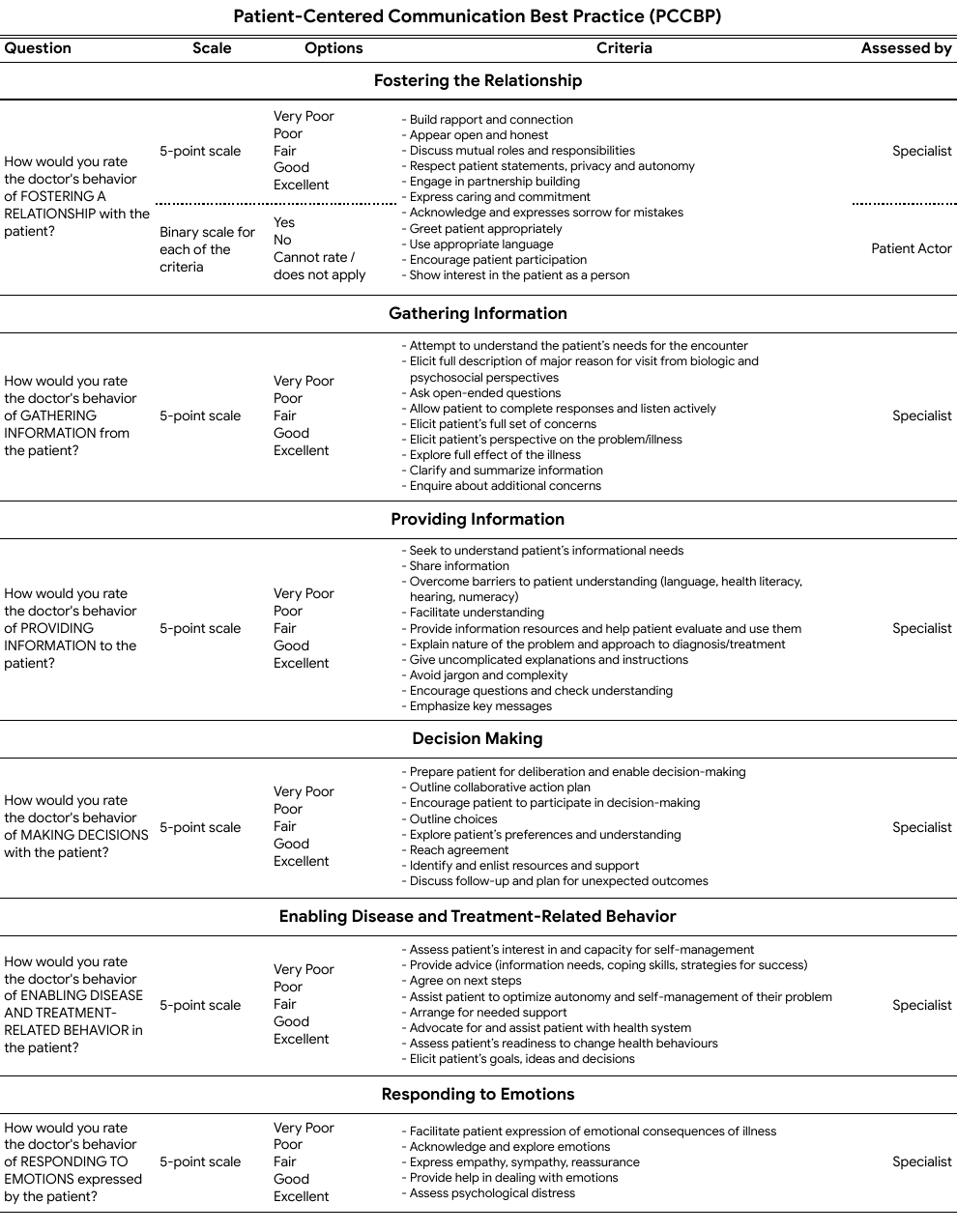}
    \caption{\textbf{Patient-Centered Communication Best Practice (PCCBP) rubric details.}}
    \label{tab:pccbp_rubric_details}
\end{table}

\begin{table}[hbt!]
    \centering
    \includegraphics[width=\textwidth,height=\textheight,keepaspectratio]{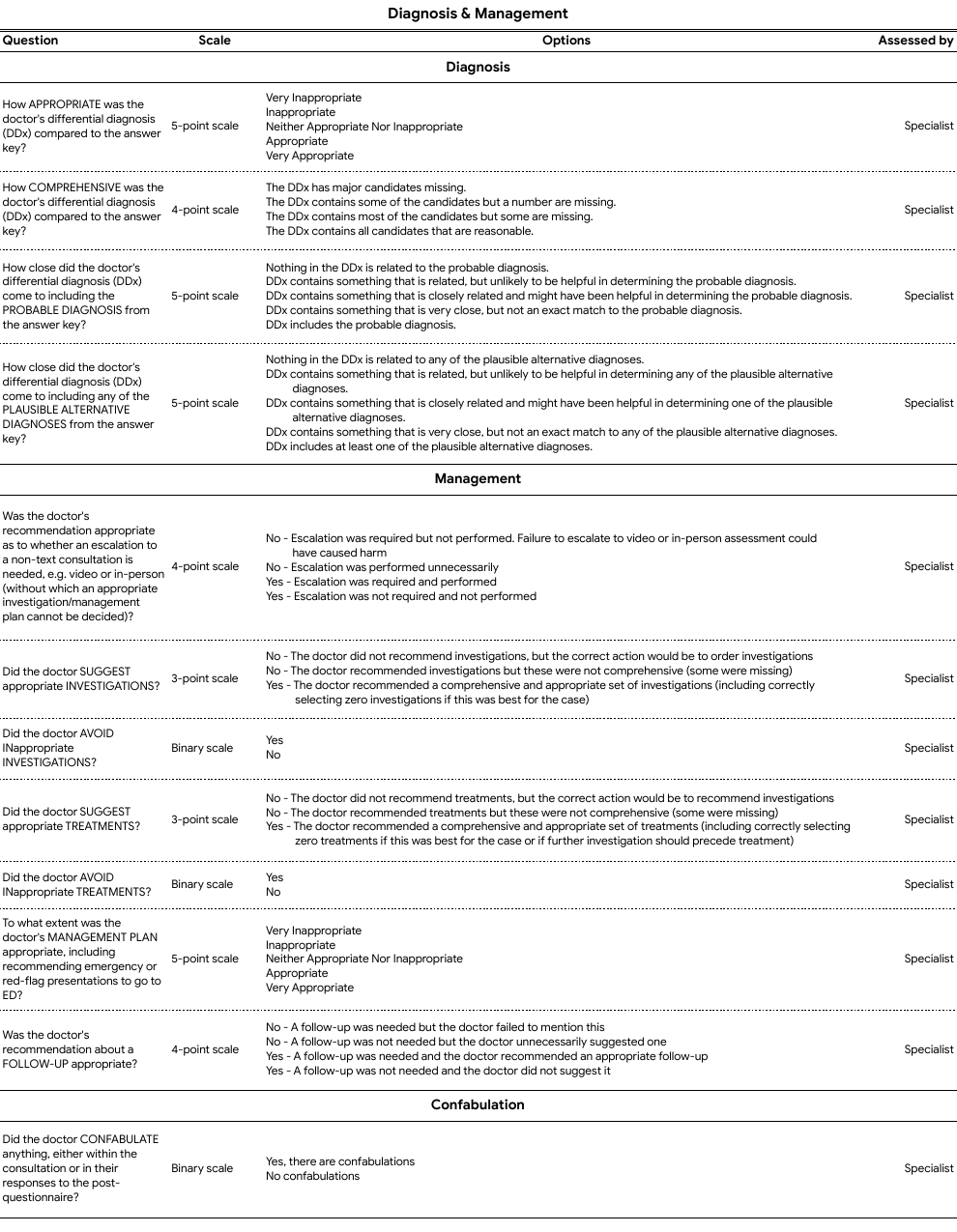}
    \caption{\textbf{Diagnosis and Management rubric details.}}
    \label{tab:diagnosis_management_rubric_details}
\end{table}

\clearpage
\section{Example of Simulated Dialogue After Self-critique}\label{synth_dialogues}

\begin{figure}[hbt!]
    \centering
    \includegraphics[width=0.95\linewidth]{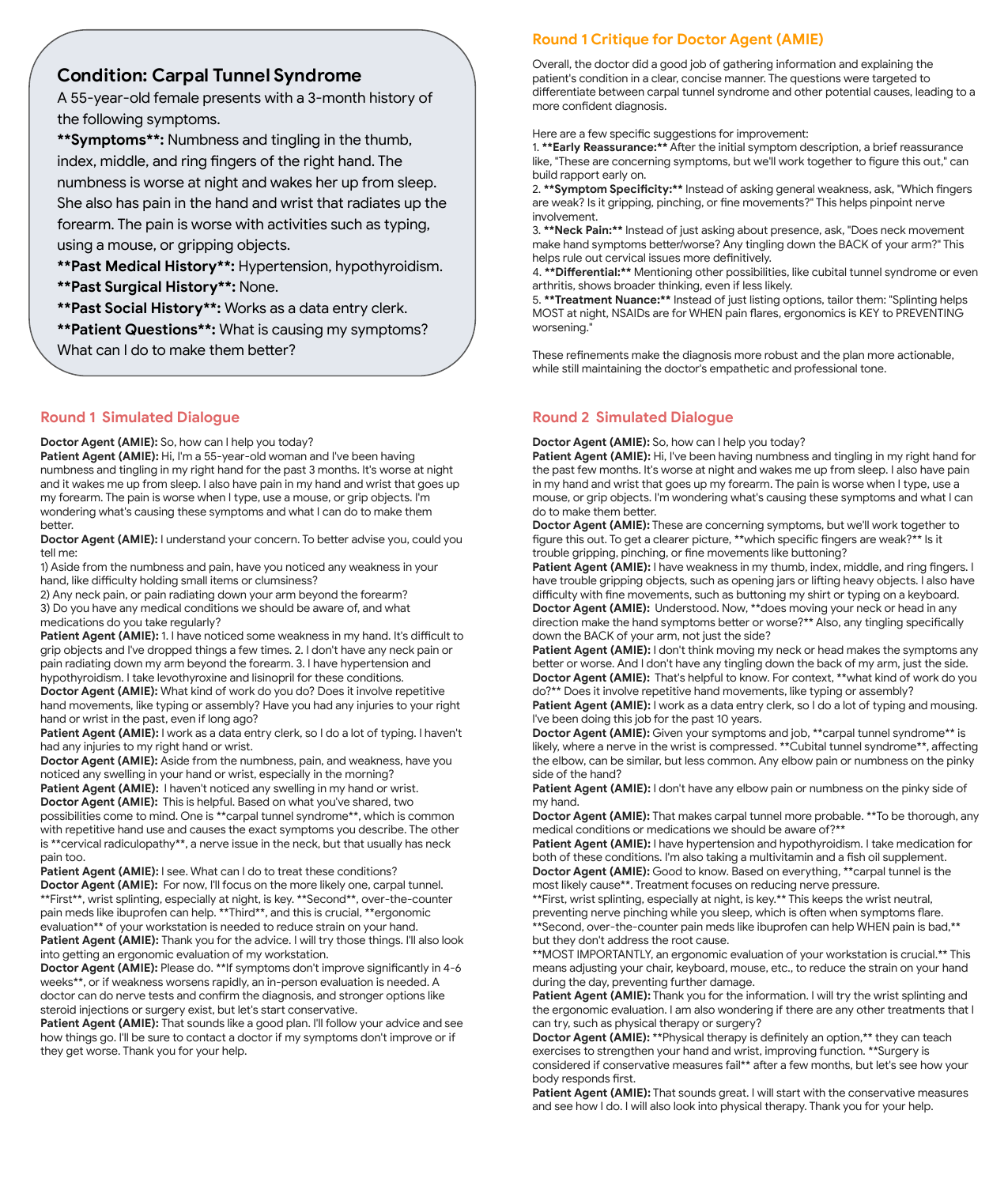}
    \caption{\textbf{Simulated Dialogue and Self-play Critique Example.} An example of AMIE modifying behavior based on in-context feedback provided during the inner-loop self-play, illustrating how this critique can impact the behavior of AMIE in the simulated dialogues. We demonstrate in \cref{fig:autoeval_selfplay} that this process on average results in improved simulated dialogue quality on a set of four PACES clinical criteria. Note that this is one preliminary example of a single round of iterative feedback and does not reflect the full simulated dialogue process. For example, in this round of feedback, the AMIE critique did not identify that management recommendations should primarily include an in-person evaluation by a physician to gauge severity and rule out more concerning causes or sequelae.}
    \label{fig:inner_loop}
\end{figure}

\clearpage
\section{AMIE User Interfaces}
\begin{figure}[hbt!]
    \centering
    \includegraphics[width=0.35\textwidth,height=\textheight,keepaspectratio]{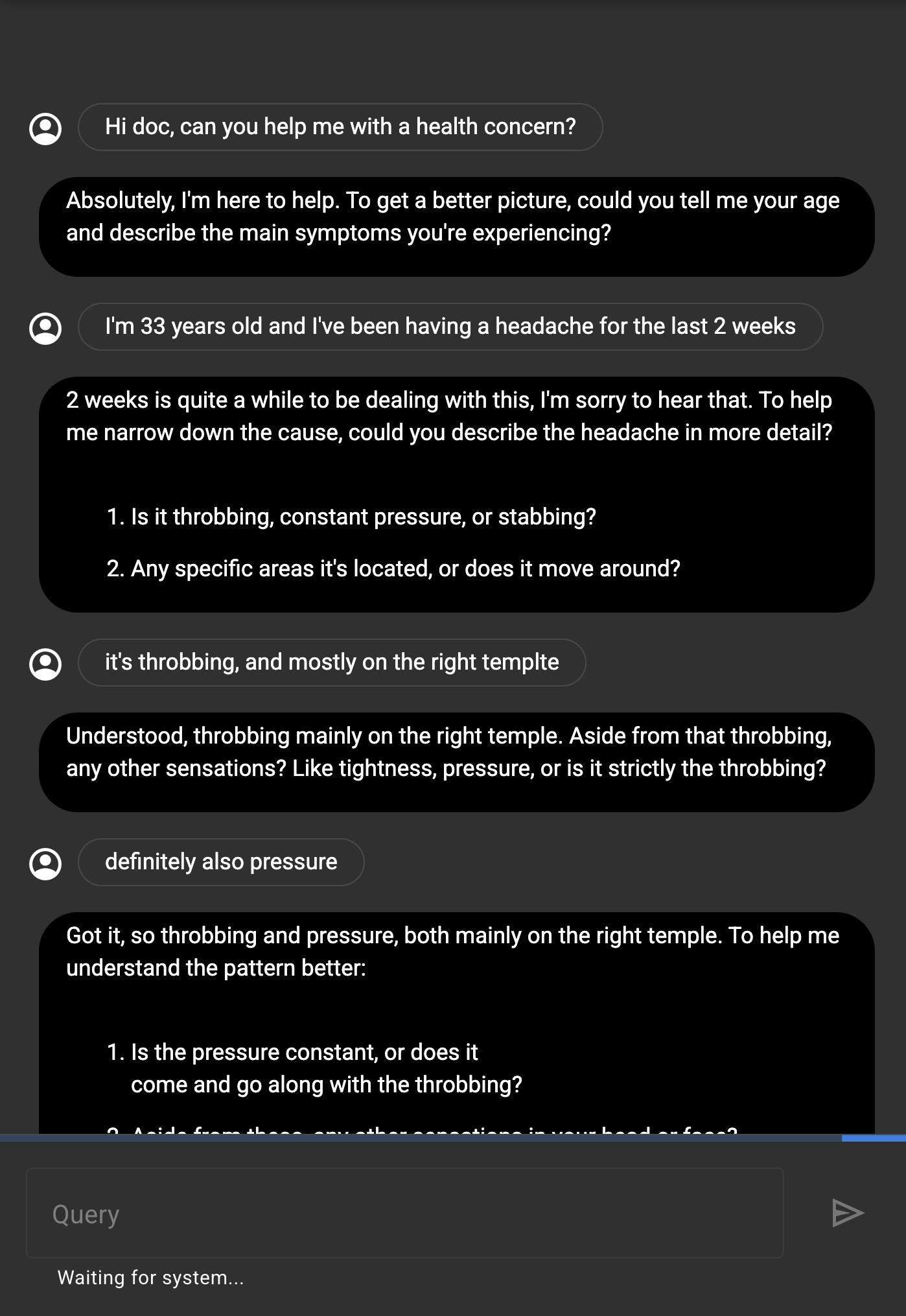}
    \vspace{0.5cm}
    \caption{\textbf{Interface for Online Text-based Consultation.}}
    \label{fig:chat_interface}
\end{figure}

\begin{figure}[hbt!]
    \centering
    \includegraphics[width=0.85\textwidth,height=\textheight,keepaspectratio]{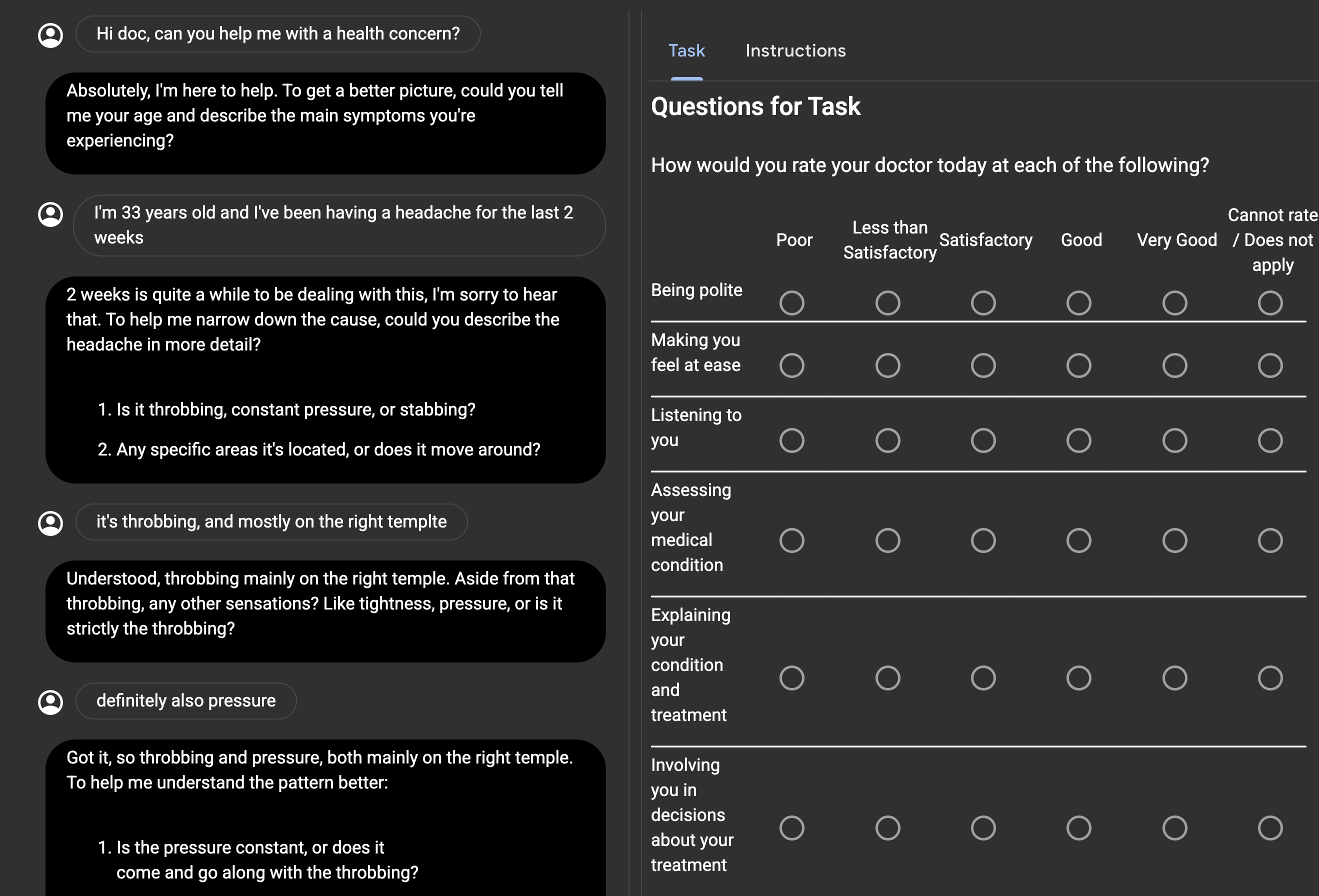}
    \vspace{0.5cm}
    \caption{\textbf{Interface for Patient Actor Ratings.}}
    \label{fig:patient_actor_rating_interface}
\end{figure}

\begin{figure}[hbt!]
    \centering
    \includegraphics[width=\textwidth,height=\textheight,keepaspectratio]{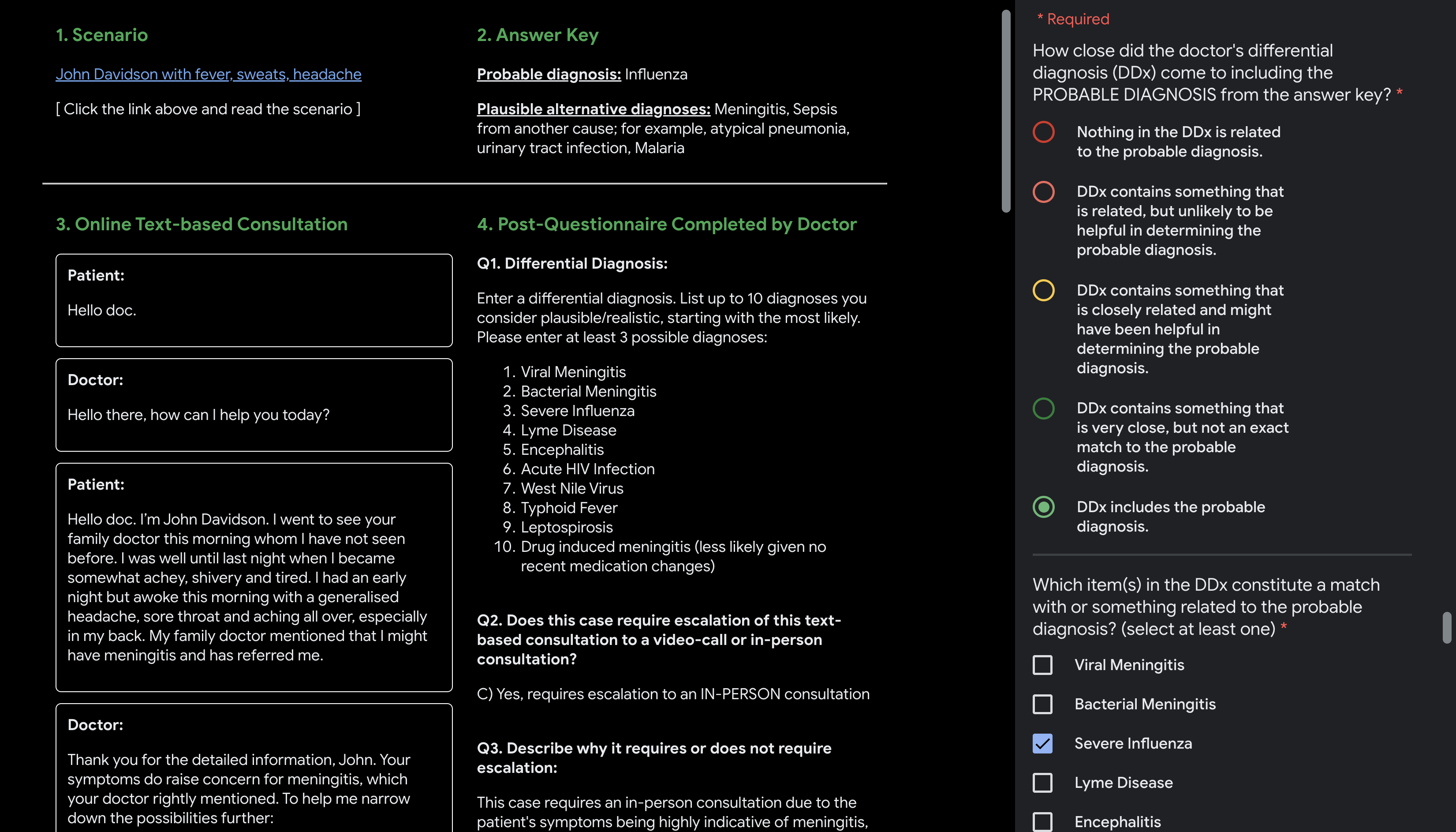}
    \vspace{0.3cm}
    \caption{\textbf{Interface for Specialist Physician Evaluation.}}
    \label{fig:specialist_evaluation_interface}
\end{figure}

\clearpage
\section{Example Consultation with OSCE Agents}
\label{appendix:osce-consultation-two-examples}

The following are two example consultations for the same scenario pack and patient actor from AMIE and PCP, respectively.

\subsection{Example AMIE Consultation}
\label{appendix:osce-consultation-amie-examples}

\begin{figure}[hbt!]
    \centering
    \includegraphics[width=0.98\textwidth,height=\textheight,keepaspectratio]{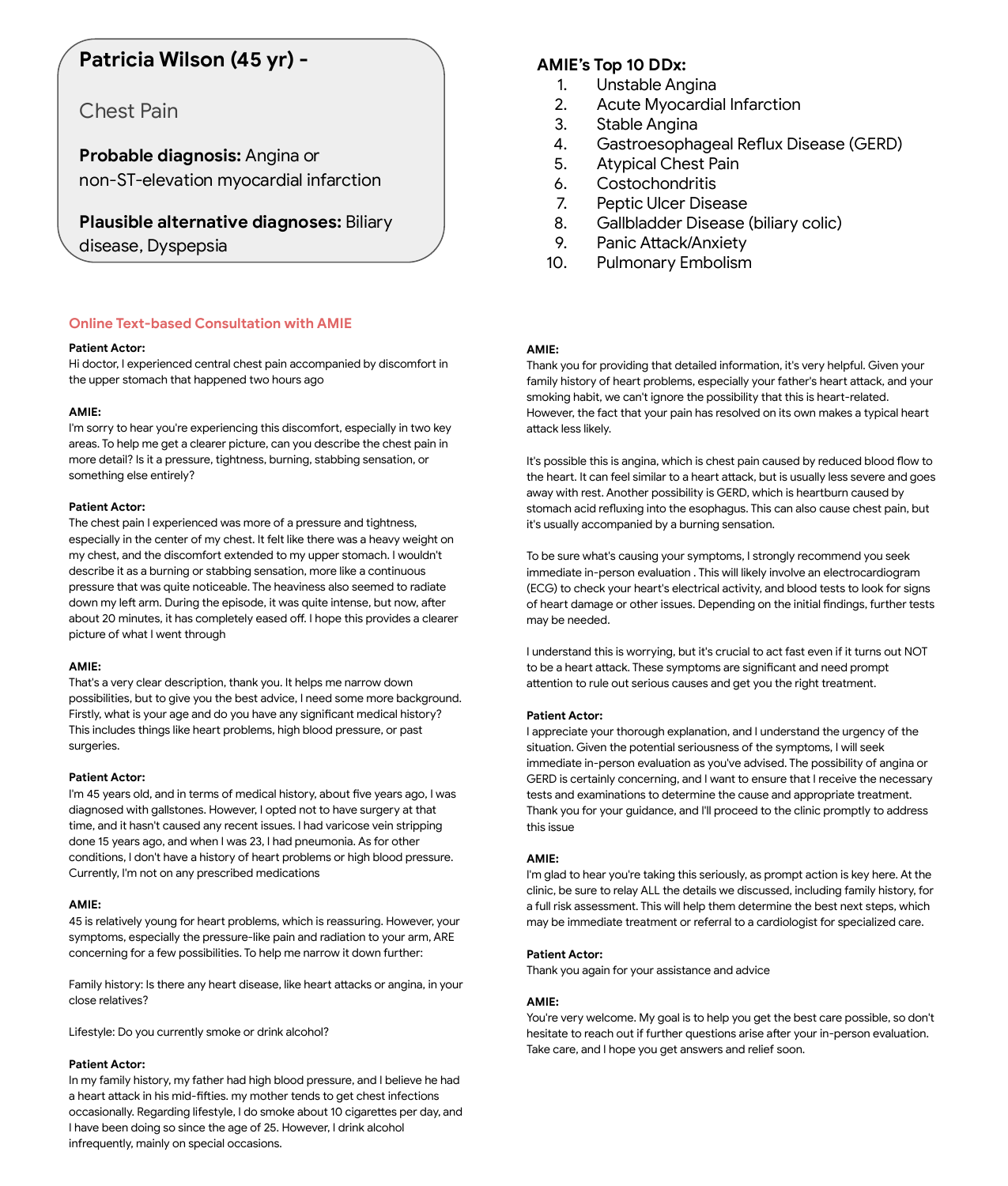}
    \vspace{0.1cm}
    \caption{\textbf{AMIE example consultation.} Example of a patient actor's conversation with AMIE during the OSCE study.}
    \label{fig:AMIE_example_osce}
\end{figure}
\clearpage
\subsection{Example PCP Consultation}
\label{appendix:osce-consultation-pcp-examples}

\begin{figure}[hbt!]
    \centering
    \includegraphics[width=\textwidth,height=\textheight,keepaspectratio]{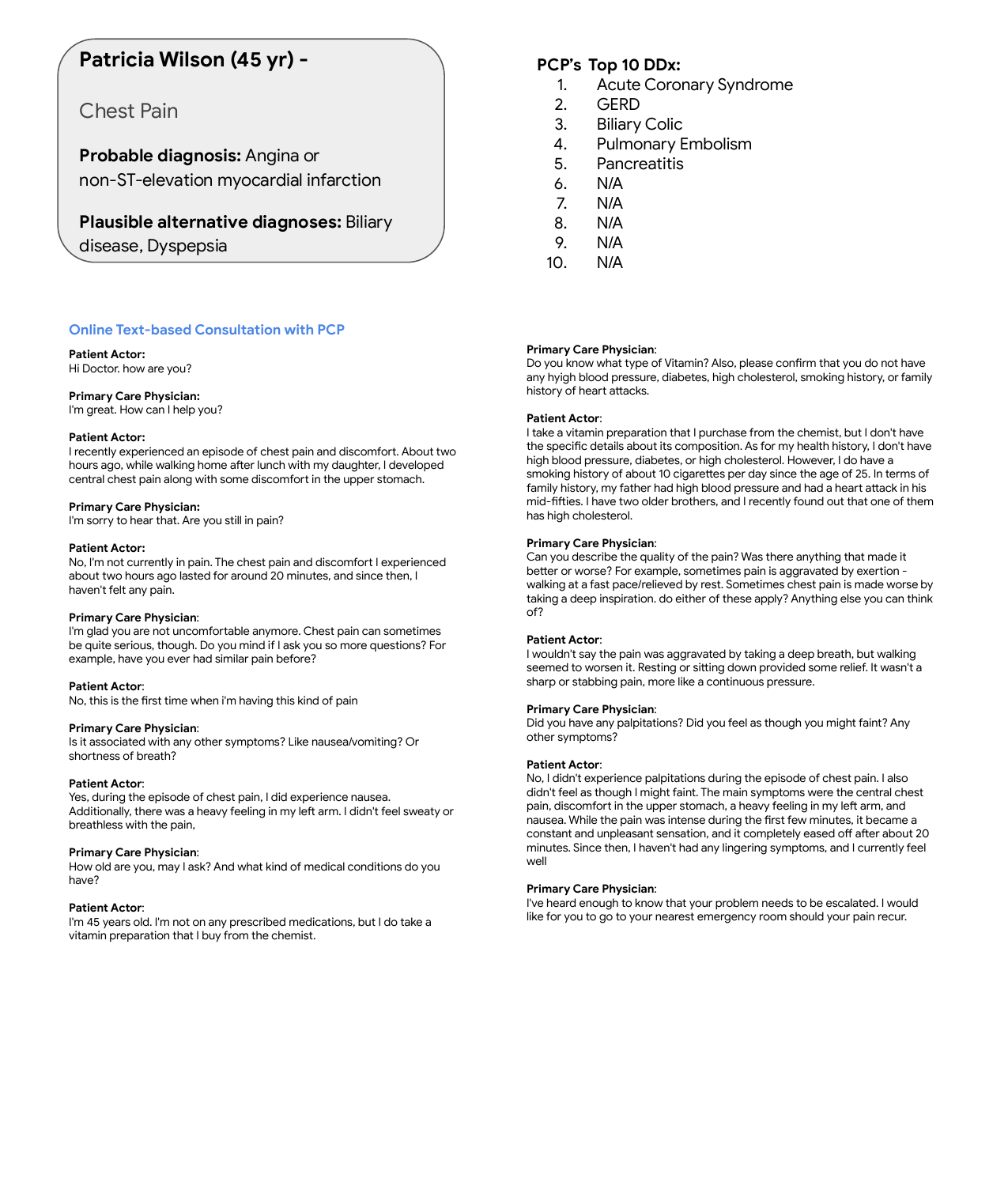}
    \vspace{-0.5cm}
    \caption{\textbf{PCP example consultation.} Example of a patient actor's conversation with a PCP during the OSCE study.}
    \label{fig:pcp_example_osce}
\end{figure}

\clearpage
\section{DDx Top-k Accuracy by Degree of Matching}
In our OSCE study, the specialist was asked the following question:
\begin{tcolorbox}
\textbf{Question}: How close did the doctor's differential diagnosis (DDx) come to including the PROBABLE DIAGNOSIS from the answer key?
\begin{itemize}
\item \textbf{(Unrelated)} Nothing in the DDx is related to the probable diagnosis.
\item \textbf{(Somewhat Related)} DDx contains something that is related, but unlikely to be helpful in determining the probable diagnosis.
\item \textbf{(Relevant)} DDx contains something that is closely related and might have been helpful in determining the probable diagnosis.
\item \textbf{(Extremely Relevant)} DDx contains something that is very close, but not an exact match to the probable diagnosis.
\item \textbf{(Exact Match)} DDx includes the probable diagnosis.
\end{itemize}
\end{tcolorbox}
\vspace{15pt}
Here we present an ablation analysis for varying degrees of matching to the ground truth where for each differential, we only considered a diagnosis a match if the specialist indicated in the answer to this question that the match was at least as close as the specified degree of matching. Note that all other specialist-rated DDx evaluations in this paper used the ``Relevant'' threshold when computing accuracy. The differences between AMIE and PCPs in DDx accuracy were statistically significant for all values of k at the matching levels ``Relevant'', ``Extremely Relevant'', and ``Exact Match''.
\vspace{15pt}
\begin{figure}[hbt!]
    \centering
    \includegraphics[width=\textwidth,height=\textheight,keepaspectratio]{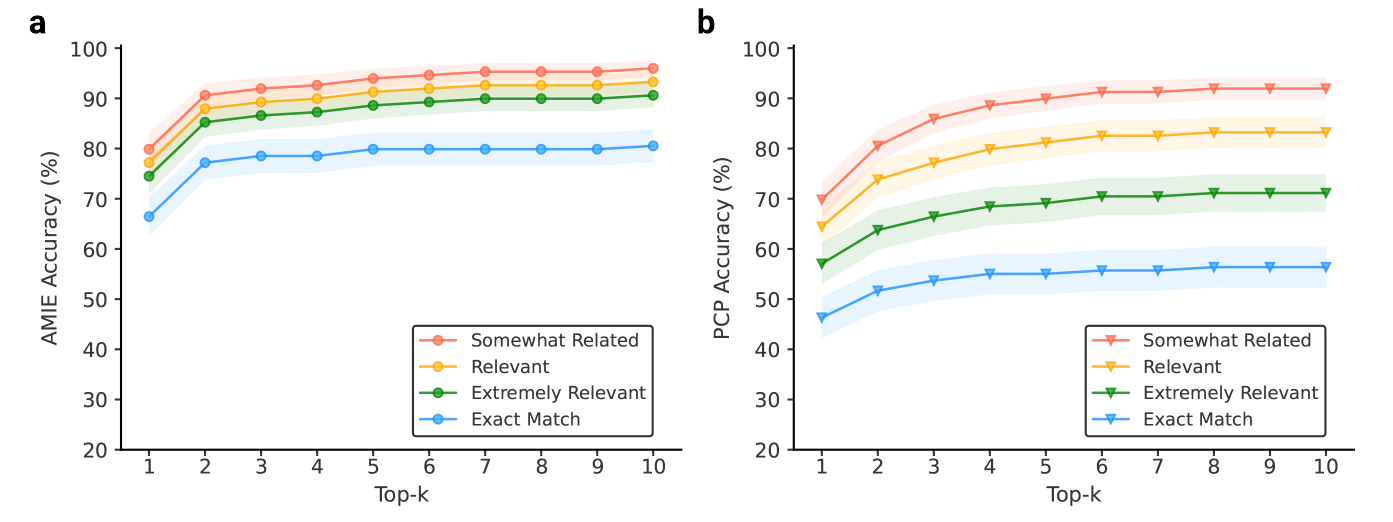}
    \caption{\textbf{Specialist rated DDx accuracy by the degree of matching.} (\textbf{a}) Specialist rated DDx top-10 accuracy for consultations conducted by AMIE. (\textbf{b}) Specialist rated DDx top-10 accuracy for consultations conducted by a PCP. For the ``Relevant'', ``Extremely Relevant'', and ``Exact Match'' levels, differences between AMIE and PCP DDx accuracy are statistically significant (bootstrap with n=10,000 and FDR correction) for all k. Differences at the ``Somewhat Related'' level are not statistically significant.}
    \label{fig:all_cases_specialist_match_cutoffs}
\end{figure}

\clearpage
\section{DDx Top-k Accuracy by Specialty}
\begin{figure}[hbt!]
    \centering
    \includegraphics[width=\textwidth,height=\textheight,keepaspectratio]{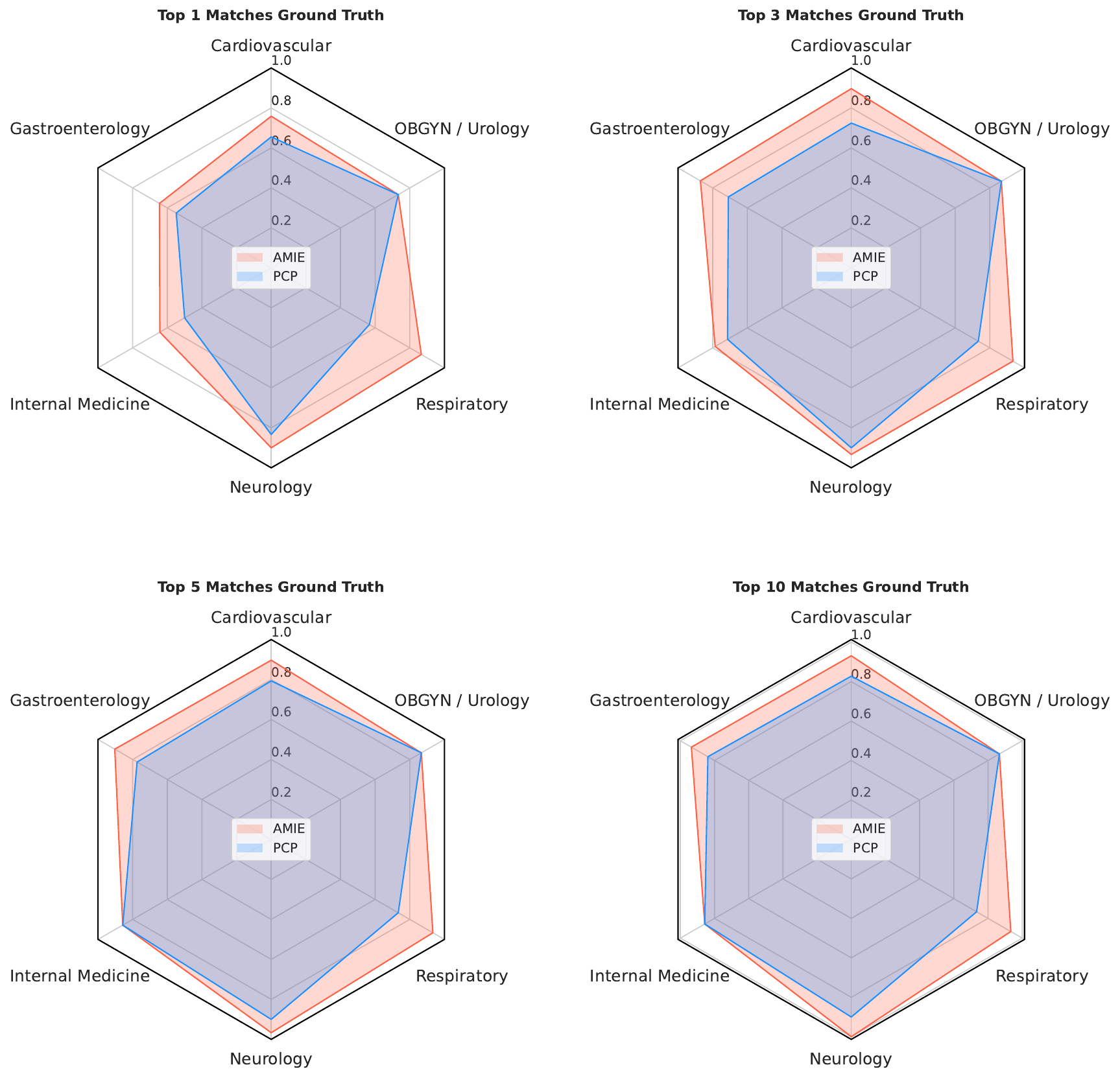}
    \caption{\textbf{Specialist rated DDx accuracy by scenario specialty.} Top 1/3/5/10 accuracy for scenarios of each specialty. Accuracies are based on the specialist ratings for AMIE and PCP differential diagnoses with respect to the ground truth. Number of dialogues per OSCE agent: Cardiology (29), Gastroenterology (31), Internal Medicine (14), Neurology (30), Respiratory (30), OBGYN / Urology (15).}
    \label{fig:specialist_ddx_ratings_by_specialty}
\end{figure}

\cref{fig:specialist_ddx_ratings_by_specialty} shows the DDx accuracy achieved by AMIE and PCPs for each specialty based on specialist ratings. Specifically, we observed that AMIE's performance matched or surpassed PCPs performance for all specialties.

\clearpage
\section{Auto-evaluation on DDx} 
\label{appendix:auto-eval-ddx}
Here we report the top-k DDx accuracy as computed by the auto-evaluation method. For each DDx in the DDx list generated by AMIE and PCPs, we used Med-PaLM 2 to determine whether the ground truth diagnosis appears within the top-k positions of the differential diagnosis list. Given a prediction and label, the auto-evaluator computes whether they match by prompting Med-PaLM 2 with the following question: \\
\begin{tcolorbox}
\textbf{DDx Auto-evaluation Prompt}\\
Is our predicted diagnosis correct (Y/N)? It is okay if the predicted diagnosis is more specific/detailed. Predicted diagnosis: \color{blue}prediction\color{black} , True diagnosis: \color{blue}label\color{black}\\Answer [Y/N]: 
\end{tcolorbox}

\subsection{Reproducing DDx Accuracy via Auto-evaluation}
\begin{figure}[hbt!]
    \centering
    \includegraphics[width=\textwidth,height=\textheight,keepaspectratio]{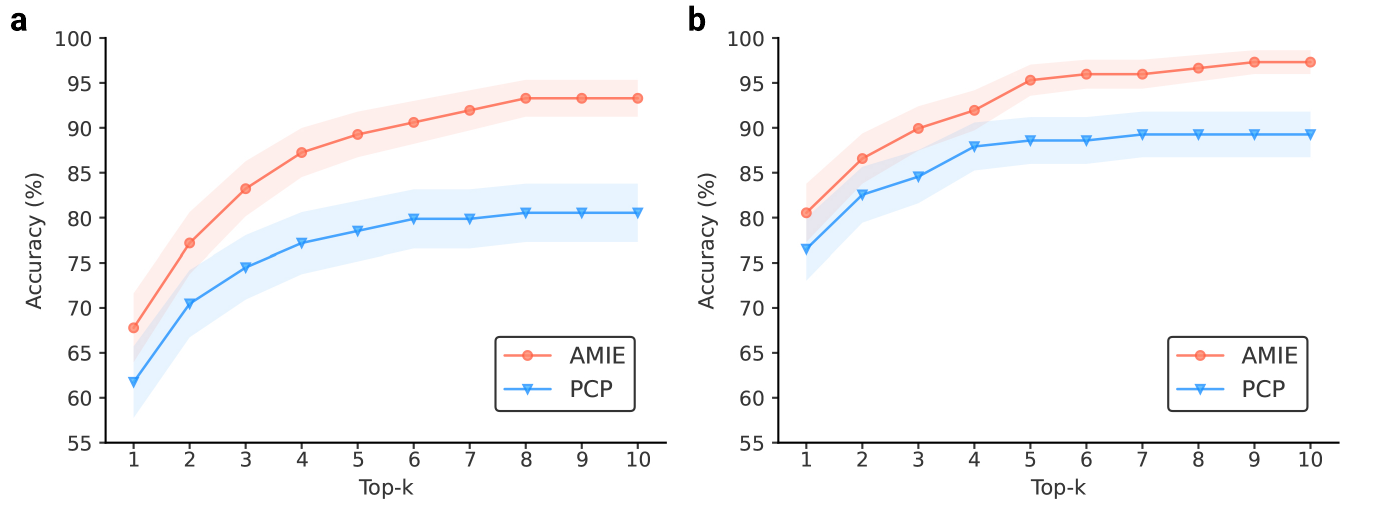}
    \caption{\textbf{Auto-evaluation rated DDx accuracy on all cases.} (\textbf{a}) Top-k auto-evaluation rating of AMIE and PCP with respect to the ground truth. Significant (with FDR correction) for $k > 2$. (\textbf{b}) Top-k auto-evaluation rating of AMIE and PCP with respect to the accepted differential. Significant (with FDR correction) for $k > 4$.}
    \label{fig:all_cases_autoeval}
\end{figure}

The overall performance trends obtained through the auto-evaluator align well with specialist assessments in~\cref{fig:all_cases_specialist_eval} despite marginal differences in the computed accuracy values, as shown in~\cref{fig:all_cases_autoeval}. These results demonstrate that the auto-evaluator is a valid surrogate for the specialist raters.

\clearpage
\subsection{AMIE DDx Accuracy on AMIE and PCP Consultations}
\begin{figure}[hbt]
    \centering
    \includegraphics[width=\textwidth,height=\textheight,keepaspectratio]{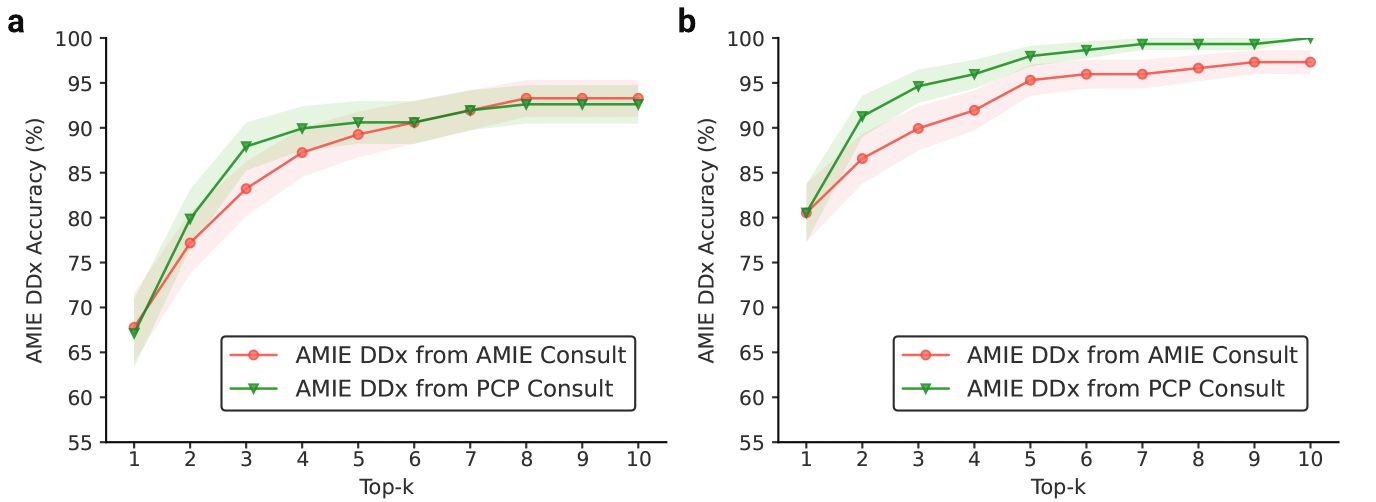}
    \caption{\textbf{Auto-evaluation rated DDx accuracy for AMIE-produced differential diagnoses from the PCP's and AMIE's consultations.} AMIE was asked to create a DDx from both the PCP's and AMIE's consultations. (\textbf{a}) Top-k auto-evaluation rating of AMIE DDx on AMIE and PCP consultations with respect to the ground truth. No differences are statistically significant. (\textbf{b}) Top-k auto-evaluation rating of AMIE DDx on AMIE and PCP consultations with respect to the accepted differential. No differences are statistically significant.}
    \label{fig:all_cases_autoeval_AMIEvsAMIE}
\end{figure}

We compared AMIE’s diagnosis accuracy based on its own consultations with its accuracy generated from corresponding PCP consultations, using the DDx auto-evaluator. Results in~\cref{fig:all_cases_autoeval_AMIEvsAMIE} showed that the diagnostic quality remained consistent regardless of whether AMIE processed information from its own dialogue or from the PCP’s conversation.

\clearpage
\subsection{DDx Accuracy as a Function of Dialogue Turns}
\begin{figure}[hbt!]
    \centering
    \includegraphics[width=\textwidth,height=\textheight,keepaspectratio]{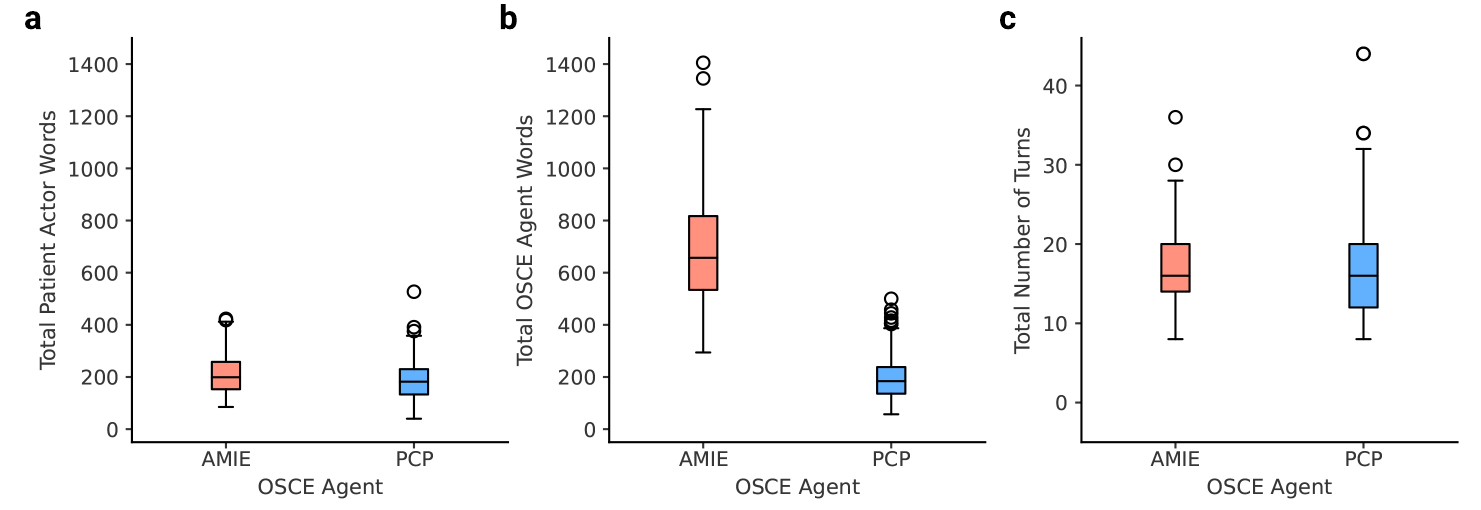}
    \caption{\textbf{Distribution of words and turns in OSCE consultations.} (\textbf{a}) Total patient actor words elicited by AMIE vs. PCPs. (\textbf{b}) Total words sent to patient actor from AMIE vs. PCPs. (\textbf{c}) Total number of turns in AMIE vs. PCP consultations.}
    \label{fig:number_of_words_and_turns}
\end{figure}

\paragraph{Distribution of words and turns.} \cref{fig:number_of_words_and_turns} shows the distributions of words and turns for the OSCE conversations. Because the number of patient actor words and turns is consistent between groups, neither agent has an unfair advantage in terms of the amount of information used to make a diagnosis. However, it is important to note that AMIE is far more verbose in its responses which may have influenced the qualitative ratings from specialists.

\begin{figure}[hbt!]
    \centering
    \includegraphics[width=\textwidth,height=\textheight,keepaspectratio]{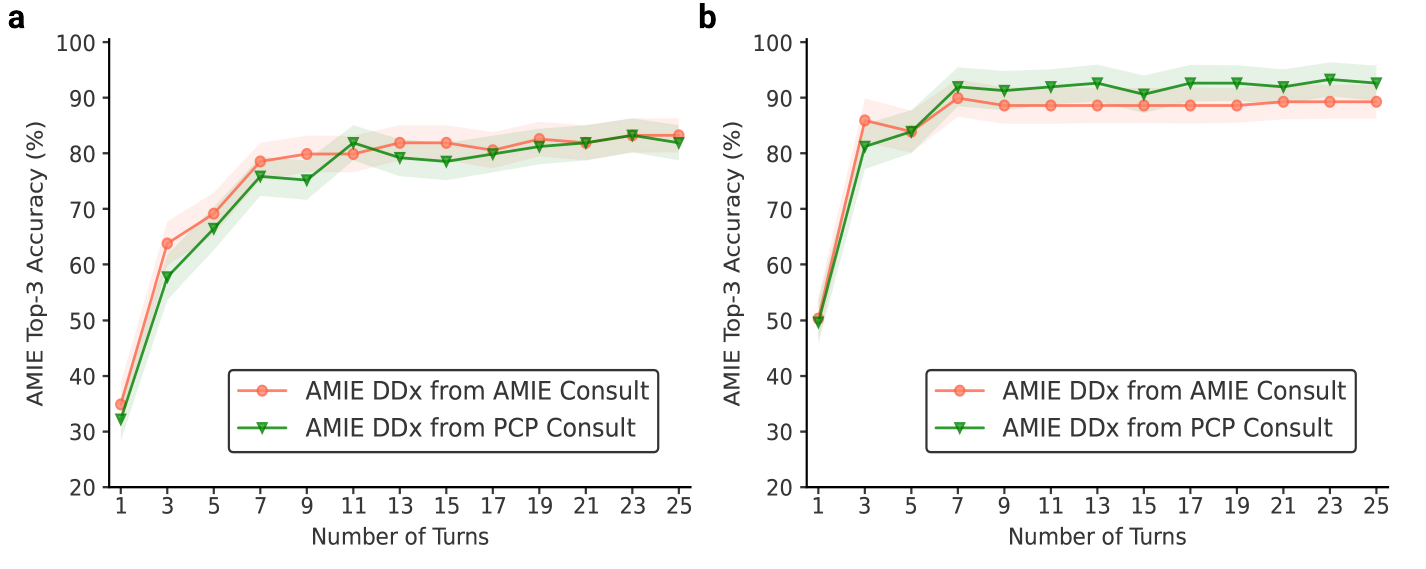}
    \caption{\textbf{Auto-evaluation rated DDx (top-3) accuracy as a function of consultation turns provided to the model.} (\textbf{a}) Top-3 auto-evaluation DDx accuracy as a function of the number of turns for the AMIE DDx on AMIE and PCP consultations with respect to the ground truth. (\textbf{b}) Top-3 auto-evaluation DDx accuracy as a function of the number of turns for the AMIE DDx on AMIE and PCP consultations with respect to the the accepted differential. No differences are statistically significant.}
    \label{fig:all_cases_autoeval_AMIEvsAMIE_turnsablation}
\end{figure}

\paragraph{Accuracy by number of turns.} Here we plotted the auto-evaluation of AMIE-generated differential diagnoses as a function of number of turns. We truncated conversations to the first $T$ turns, and then asked AMIE to produce a DDx with this truncated conversation. For both the AMIE and PCP conversations, we observed that AMIE's average diagnostic accuracy began to plateau within 10 turns, with additional information gathering having diminishing returns on the diagnostic performance as shown in~\cref{fig:all_cases_autoeval_AMIEvsAMIE_turnsablation}.

\clearpage
\section{DDx Accuracy by Location}

\paragraph{Accuracy by Location.} We compared the specialist ratings for the 67 scenarios conducted in Canada and the 82 scenarios conducted in India.

\begin{figure}[hbt!]
    \centering
    \includegraphics[width=\textwidth,height=\textheight,keepaspectratio]{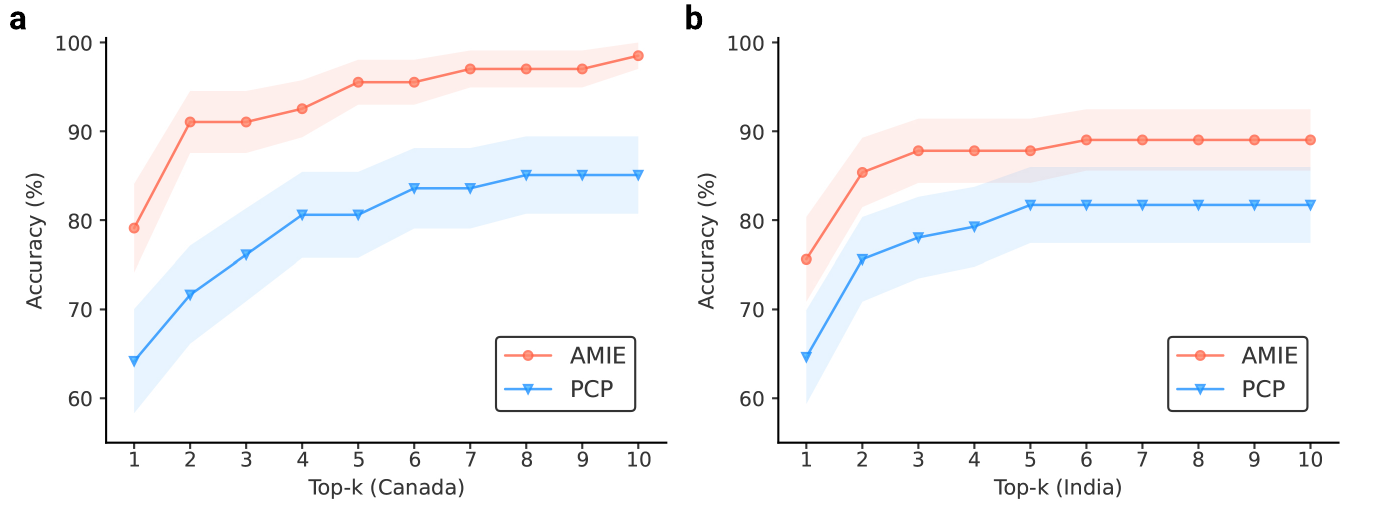}
    \caption{\textbf{Specialist rated DDx accuracy by location.} (\textbf{a}) Specialist DDx rating of AMIE and PCP with respect to the ground truth for the 67 cases conducted in Canada. Accuracies at all k positions are significant with FDR correction. (\textbf{b}) Specialist DDx rating of AMIE and PCP with respect to the ground truth for the 82 cases conducted in India. While the trends are the same as in Canada, the differences between AMIE and PCP are not statistically significant with FDR correction.}
    \label{fig:per_location_specialist_eval}
\end{figure}

\paragraph{Shared Scenarios.} We repeated 40 of the scenarios at the other location, meaning if it was originally run in Canada, we then ran it in India and vice-versa. This included all of the UK scenarios and 26 of the India scenarios. Because these conversations did not have specialist ratings, we instead leveraged auto-evaluation to compare the produced differential diagnoses and ablate the effect of the OSCE location.

\begin{figure}[hbt!]
   \centering
    \includegraphics[width=\textwidth,height=\textheight,keepaspectratio]{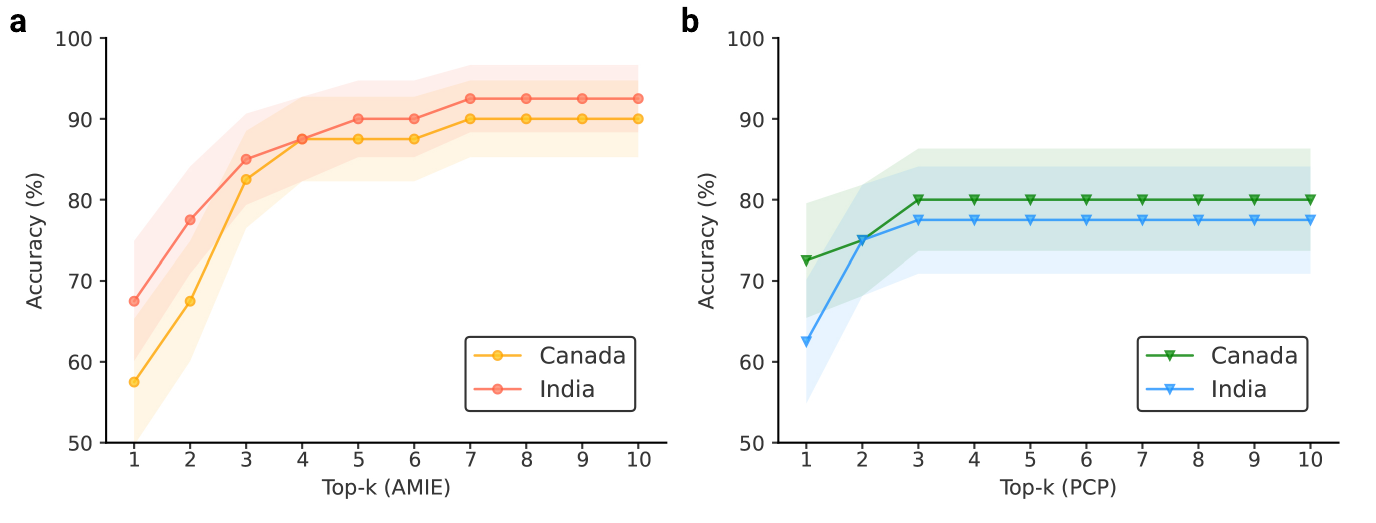}
    \caption{\textbf{Auto-evaluation rated DDx accuracy for scenarios conducted at both testing locations.} (\textbf{a}) Auto-evaluation rated top-k DDx performance of AMIE on a set of 40 scenarios conducted in both locations. (\textbf{b}) Auto-evaluation rated top-k DDx performance of the PCPs on a set of 40 scenarios conducted in both locations.}
    \label{fig:same_scenario_location_ddx}
\end{figure}

\paragraph{Results.} We observed a higher average diagnostic performance for AMIE in Canada than in India (see~\cref{fig:per_location_specialist_eval}). However, when comparing scenarios performed at both study locations, we observed that AMIE and PCP performance remained consistent regardless of the study location (see~\cref{fig:same_scenario_location_ddx}), suggesting that the observed performance variations are likely not due to the patient actor or clinician rater differences, but instead might be attributed to inherent differences in the difficulty levels of scenarios in each location.

\clearpage
\section{Model-based Auto-evaluation of Qualitative Criteria}
\label{appendix:auto-eval}
In order to accurately emulate the ratings of specialists on the clinical criteria in our OSCE evaluation framework, we developed a model-based auto-evaluation procedure leveraging the AMIE model to score dialogues from 1 to 5 based on how well they exemplified those qualitative criteria. We initially focused on a subset of four clinical axes from the PACES criteria (see \cref{tab:paces_rubric_details}), however, this procedure can be easily extended to other ratings.

\begin{figure}[hbt!]
    \centering
    \includegraphics[width=0.95\linewidth]{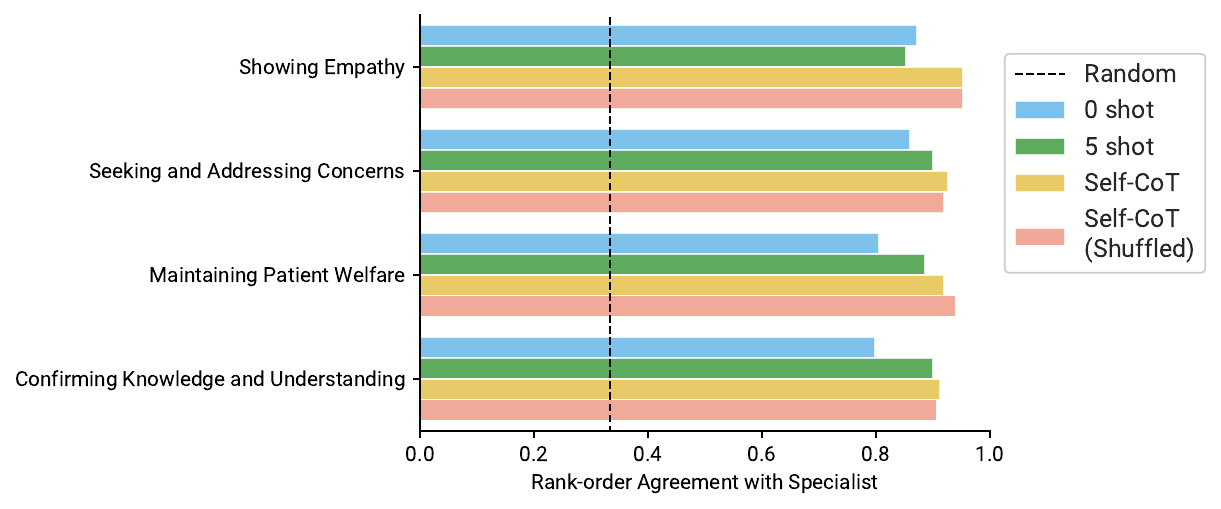}
    \caption{\textbf{Rank-order agreement to specialist ratings of all 149 dialogue pairs, comparing various auto-evaluation prompting techniques.} We choose to leverage the self-CoT technique for the auto-evaluation of clinical criteria.}
    \label{fig:autoeval_ablation}
\end{figure}

\begin{figure}[hbt!]
    \centering
    \includegraphics[width=1\linewidth]{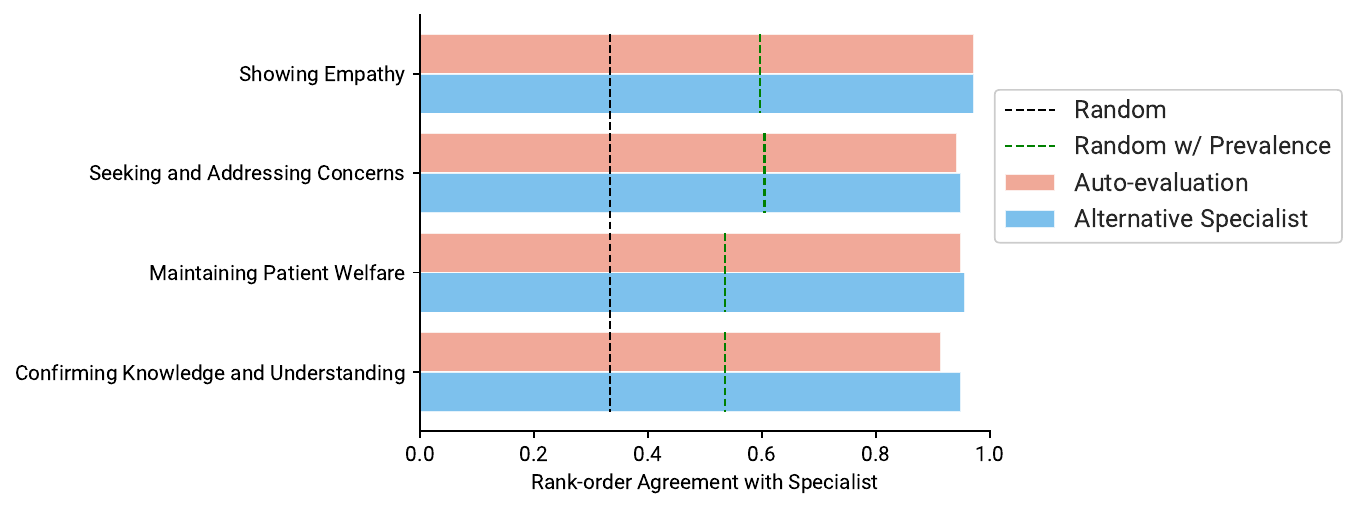}
    \caption{\textbf{Rank-order agreement to specialist ratings of 139 dialogue pairs (excluding cases without multiple specialist ratings) for alternative specialists compared to the self-CoT auto-evaluation technique.} Auto-evaluation agreement to the first specialist is comparable to inter-specialist agreement. The black dashed line shows the rank-order agreement one would get with a random ranking of the AMIE and PCP dialogues, while the green dashed line shows the rank-order agreement with a strategy of randomly guessing according to the distribution of specialist preferences for each criteria.}
    \label{fig:autoeval_vs_specialist}
\end{figure}

\begin{figure}[hbt!]
    \centering
    \includegraphics[width=0.9\linewidth,height=\textheight,keepaspectratio]{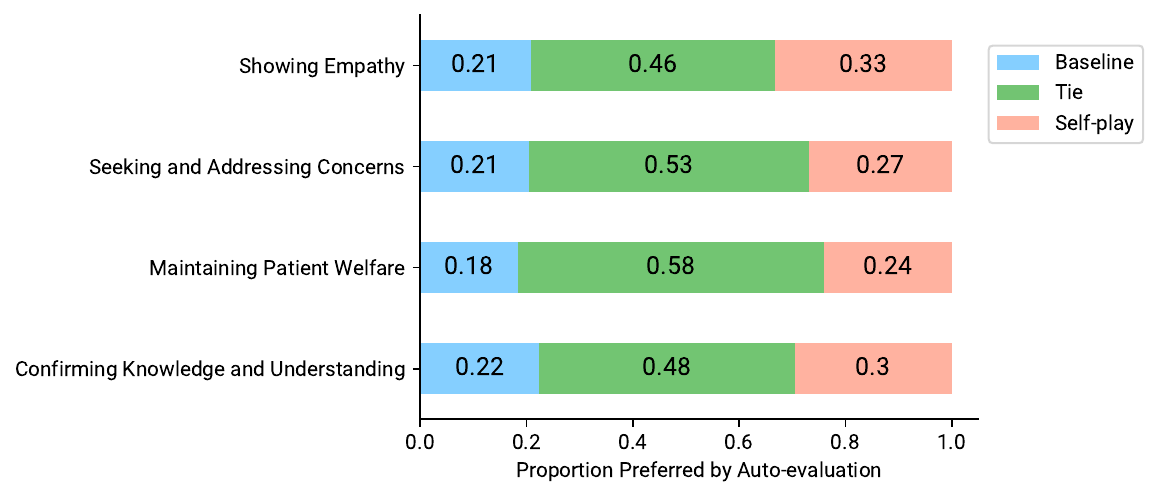}
    \caption{\textbf{Comparison of the simulated dialogue quality before and after self-play as assessed by auto-evaluation.} The self-play dialogues after one round of critique are preferred by auto-evaluation more often than the baseline dialogue generated without revision.}
    \label{fig:autoeval_selfplay}
\end{figure}

Using the 298 dialogues produced by AMIE and PCPs in this study, we corroborated the results of the specialist ratings on these 4 criteria using the auto-evaluation procedure. We validated that the auto-evaluation rankings were well aligned with these specialist ratings (see \cref{fig:autoeval_ablation,fig:autoeval_vs_specialist}). Additionally, we applied it to the simulated dialogues generated via the inner-loop self-play procedure to test whether this iterative process resulted in measurable improvements in dialogue quality (see \cref{fig:autoeval_selfplay}).

\paragraph{Self-CoT Procedure for Auto-Evaluation of Clinical Criteria.}
The auto-evaluation procedure we employed was a two-step process in which we prompted AMIE itself to rate dialogues on the chosen subset of the PACES criteria (see~\cref{tab:paces_rubric_details}).
\begin{enumerate}
\item First, we prompted AMIE to summarize good and bad aspects of several dialogues and provide an explanation of the provided human rating between 1 and 5 (see \cref{box:autoeval-explanation-prompt}). 
\item Next, we used these self-generated explanations alongside their respective dialogues as examples in a 5-shot prompt to evaluate and rate a new dialogue. This few-shot prompt included one example for each point on the 5-point rating scale (see  \cref{box:autoeval-rating-prompt}).
\end{enumerate}
In both prompts, we included the rating scale and expert-derived examples of good or bad behaviour for a particular criterion, matching those shown in \cref{tab:paces_rubric_details}. We referred to this prompting method as self-CoT (Chain-of-Thoughts)~\cite{wei2022chain} as the plausible reasoning for the human ratings are derived from the model itself. 

\paragraph{Rank-order Agreement.} We evaluated our auto-evaluation method by quantifying its agreement with the specialist rankings of the OSCE dialogues. We limited our analysis to the 149 dialogue pairs in the study. Thus, each pair consisted of a AMIE conversation and a PCP conversation with the same patient actor, and rated by the same specialist. For a pair of two dialogues, the three possibilities were: the first one was rated better than the second one, they were equally rated, or the first one was rated worse than the second one. We defined the rank-order agreement as the proportion of dialogue pairs for which the specialist ranking was preserved by the auto-evaluation ratings. For example, we counted it as correct when our auto-evaluation rated AMIE's dialogue as better than the PCP's dialogue if specialists also rated AMIE's dialogue as better, regardless of the exact scores each method assigned.

\paragraph{Auto-evaluation Prompting Strategies.} Using the rank-order agreement metric, we ablated the effect of the two-step prompting and compared it to other methods such as the five-shot prompting (i.e. dropping step 1), shuffled five-shot self-CoT prompting where the order of support examples was randomised each time, and 0-shot prompting using only the rating scale explanation itself (see \cref{fig:autoeval_ablation}). All methods outperformed the chance level, with the two-step process generally outperforming other methods, though this difference was marginal. Shuffling the examples in the self-CoT prompt made no difference on average. 

\paragraph{Benchmarking Auto-evaluation.} While auto-evaluation was significantly better than random guessing at aligning with specialist preferences, it was unclear if the resulting performance was sufficient. To test this, we had 139 dialogue pairs each rated by a second specialist, and we computed the rank-order agreement of the two specialists on this subset (see figure \ref{fig:autoeval_vs_specialist}). We observed that auto-evaluation was about as accurate as the alternative specialists in predicting the first specialist's rank-order, suggesting that it is useful to leverage auto-evaluation for these criteria.

\paragraph{Evaluating Self-play Dialogues.} We applied our auto-evaluation procedure to 1,142 dialogues (derived from common conditions) before and after being refined through the self-play critique. We demonstrated that, on average, the refined dialogues after an iteration of critique/revision were rated higher than the original baseline dialogues across all criteria \ref{fig:autoeval_selfplay}.

\begin{figure}[hbt!]
\caption{\textbf{Prompt for the explanation generation given an dialogue and the human rating.} AMIE was prompted to summarize good and bad aspects of a particular dialogue and provide an explanation of the given human rating between 1 and 5.}
\label{box:autoeval-explanation-prompt}
\vspace{0.1cm}
\begin{tcolorbox}[title=Prompt for Explanation Generation for Dialogue and Human Rating]
    I have a doctor-patient dialogue and the corresponding rating that quantifies its quality according to the following criterion: <criterion> (e.g., maintaining patient welfare). The rating of the dialogue is on a scale of 1 to 5 where: \\
    \\
    5: <definition> e.g., ``Treats patient respectfully, and ensures comfort, safety and dignity'' \\
    1: <definition> e.g., ``Causes patient physical or emotional discomfort AND jeopardises patient safety'' \\
    \\
    First, describe which parts of the dialogue are good with respect to the criterion.
    Then, describe which parts are bad with respect to the criterion. Lastly, summarise the above to explain the provided rating, using the following format: \\ 
    \\
    Good: ... \\
    Bad: ... \\
    Summary: ... \\ 
    \\
    DIALOGUE: <dialogue> \\
    Rating: <human rating> \\
    EVALUATION: \\
    \\
    \textbf{Example output (for a dialogue with rating 4 on `maintaining patient welfare'):}\\
    Good: The doctor took the patient's concerns seriously and acted quickly to address the situation. They asked pertinent questions to gather information about the patient's symptoms and medical history. They also provided clear instructions on what the patient needed to do next.\\
    \\
    Bad: The doctor did not provide much emotional support to the patient. They could have been more reassuring and empathetic towards the patient's fear and anxiety. \\
    \\
    Summary: Overall, the doctor did a good job of maintaining patient welfare in this situation. They took prompt action to address the patient's medical emergency and provided clear instructions to the patient. However, they could have been more attentive to the patient's emotional needs.
\end{tcolorbox}
\end{figure}

\begin{figure}[hbt!]
\caption{\textbf{Prompt for the auto-evaluation of qualitative criteria.} We constructed few-shot prompt with self-generated explanations alongside their respective dialogues as exemplars to evaluate and rate a new dialogue. This few-shot prompt included one example for each point on the 5-point rating scale.}
\label{box:autoeval-rating-prompt}
\vspace{0.1cm}
\begin{tcolorbox}[title=Prompt for Auto-Evaluation]
    I have a doctor-patient dialogue which I would like you to evaluate on the following criterion: <criterion> (e.g., maintaining patient welfare). The dialogue should be rated on a scale of 1-5 with respect to the criterion where: \\
    \\
    5: <definition> e.g., ``Treats patient respectfully, and ensures comfort, safety and dignity'' \\
    1: <definition> e.g., ``Causes patient physical or emotional discomfort AND jeopardises patient safety'' \\
    \\
    Here are some example dialogues and their ratings: \\
    DIALOGUE: <example dialog> \\
    EVALUATION: <example self-generated explanation> \\
    Rating: <example rating> \\ 
    ... \\
    
    Now, please rate the following dialogue as instructed below. First, describe which parts of the dialogue are good with respect to the criterion. Then, describe which parts are bad with respect to the criterion. Third, summarise the above findings. Lastly, rate the dialogue on a scale of 1-5 with respect to the criterion, according to this schema: \\ 
    \\
    Good: ... \\
    Bad: ... \\
    Summary: ... \\ 
    Rating: ... \\ 
    \\
    DIALOGUE: <dialogue> \\
    EVALUATION: \\
\end{tcolorbox}
\end{figure}

\clearpage
\subsection{Rank-order Agreement of Auto-evaluation to Specialist}

\clearpage
\subsection{Auto-evaluation of Simulated Dialogues with Self-play}

\clearpage

\begin{table}[]
\centering
\small
\caption{Model card for AMIE.}
\vspace{3pt}
\label{tab:model-card-med-palm}
\begin{tabular}{lp{0.6\textwidth}}
\toprule
\textbf{}            & \textbf{Model characteristics}                               \\ 
Model initialization & The model was initialized from an LLM similar to PaLM-2 \cite{google2023palm2}. Additional fine-tuning was performed as described in~\cref{sec:instruction-finetuning}. \\

                     & \textbf{Usage}                                               \\
Application &
  The primary use is research on LLMs for clinical history-taking and medical dialogue including advancing diagnostic accuracy, alignment methods, fairness, safety, and equity research, and understanding limitations of current LLMs for such applications. \\ \hline
                     & \textbf{Data overview}                                       \\
Training dataset &
  The dataset was datasets of medical question answering/reasoning, summarization and medical dialogue datasets in~\cref{sec:datasets}. \\
Evaluation &
Randomized OSCE was performed to understand AMIE's capabilities and benchmark against expert primary care physicians. \\
                     & \textbf{Evaluation results}                                  \\
Evaluation results   & The results are described in~\cref{sec:results}.          \\
\bottomrule
\end{tabular}
\end{table}


\newpage
\setlength\bibitemsep{3pt}
\printbibliography
\balance
\clearpage
\end{refsection}

\end{document}